%% file: main.tex
\documentclass[acmtog,nonacm]{acmart}

\title{A comprehensive study of time-of-flight non-line-of-sight imaging}
\author{Julio Marco}
\authornote{Corresponding author, juliom@unizar.es}
\affiliation{%
  \institution{Universidad de Zaragoza--I3A}
  \country{Spain}
}
\author{Adrian Jarabo}
\affiliation{%
  \institution{Universidad de Zaragoza--I3A}
  \country{Spain}
}
\author{Ji Hyun Nam}
\affiliation{%
  \institution{University of Wisconsin--Madison}
  \country{USA}
}
\affiliation{%
  \institution{Department of Data Science, Inha University}
  \country{South Korea}
}
\author{Alberto Tosi}
\affiliation{%
  \institution{Politecnico di Milano}
  \country{Italy}
}
\author{Diego Gutierrez}
\affiliation{%
  \institution{Universidad de Zaragoza--I3A}
  \country{Spain}
}

\author{Andreas Velten}
\affiliation{%
  \institution{University of Wisconsin--Madison}
  \country{USA}
}

\usepackage{epsfig}
\usepackage{graphicx}

\usepackage[dvipsnames]{xcolor}

\usepackage{soul}
\usepackage[many]{tcolorbox}

\usepackage{pstricks}
\usepackage{graphicx}
\usepackage{caption}
\usepackage{subcaption}
\usepackage{xspace}
\usepackage{cancel}
\usepackage{mathtools}
\usepackage{booktabs}
\usepackage{siunitx}
\usepackage[inline]{enumitem}
\usepackage{graphicx}
\usepackage{tabularx}
\usepackage{animate}
\usepackage{empheq}
\usepackage{commath}
\usepackage{stmaryrd}
\usepackage{relsize}
\usepackage{svg} 
\usepackage{makecell}
\usepackage{resizegather}
\usepackage[capitalise]{cleveref}
\usepackage{hyperref}

\PassOptionsToPackage{doi=false}{natbib}

\graphicspath{ {./figures/}}

\input{src_flags}
\DataAvailabilityfalse
\TOGtrue
\input{src_defines}
\authorsaddresses{}

\begin{document}
\begin{abstract}
    \input{src0_abstract}
\end{abstract}
\maketitle

\input{src1_intro}
\input{src2_related_work}
\input{src3_problem_statement}

\input{src4_NLOS_inverses}
\input{src5_frequencydomain}
\input{src6_sim_experiments}
\input{src7_real_experiments}
\input{src8_conclusions}
{
\bibliographystyle{ACM-Reference-Format}
\bibliography{bibliography}
}
\end{document}

%% file: src_flags.tex
\newif\ifDataAvailability
\newif\ifTOG
\newif\ifTPAMI

%% file: src_defines.tex
\newcommand{\erref}[2]{{Eqs.~(\ref{#1}) to (\ref{#2})}}
\newcommand{\eeeref}[3]{{Eqs.~(\ref{#1}, (\ref{#2}), and (\ref{#3})}}

\ifTPAMI
\renewcommand{\paragraph}[1]{\textbf{#1}.\quad}
\fi

\definecolor{red}{rgb}{0.8,0,0}
\definecolor{lightcyan}{rgb}{0.7,1.0,1.0}
\definecolor{lightyellow}{rgb}{1.0,1.0,0.7}
\definecolor{purered}{rgb}{1,0,0}
\definecolor{magenta}{rgb}{1,0,1}
\definecolor{darkred}{rgb}{0.6,0,0}
\definecolor{green}{rgb}{0.0,0.5,0}
\definecolor{blue}{rgb}{0,0,1}
\definecolor{paleblue}{rgb}{0.5,0.5,1}
\definecolor{darkblue}{rgb}{0,0,0.55}
\definecolor{orange}{rgb}{0.9,0.3,0.1}
\definecolor{purple}{rgb}{0.6,0.0,0.6}
\definecolor{cyan}{rgb}{0.0,0.7,0.7}
\definecolor{darkgray}{rgb}{0.4,0.4,0.4}
\definecolor{bronze}{rgb}{0.8, 0.5, 0.2}
\definecolor{dorange}{rgb}{0.75, 0.4, 0.0}
\ifTOG
\newtcolorbox{eqboxedwhite}[1][]{colback=white,size=small,width=\linewidth, boxsep=2pt, bottom=0.2em, top=-0.5em, after=}
\else
\newtcolorbox{eqboxedwhite}[1][]{colback=white,size=small,width=\linewidth, boxsep=2pt, bottom=1pt, top=-1em, after=}
\fi

\newcommand{\comm}[3]{\leavevmode\textcolor{#1}{\emph{(\textbf{#2:} #3)}}}
\renewcommand{\comm}[3]{}

\newcommand{\transpose}{{\scriptstyle\top}}
\renewcommand{\transpose}{{\intercal}}
\newcommand{\minv}[1]{{#1}^{-1}}
\newcommand{\Fspace}[1]{\widehat{#1}}

\newcommand{\mat}[1]{{\mathbf{#1}}}
\newcommand{\reparam}[1]{\mathrm{#1}}
\newcommand{\freq}{\Omega}
\newcommand\given[1][]{\:#1\vert\:}
\newcommand{\conv}{\ast}

\newcommand{\sol}{c}
\newcommand{\diff}{\mathrm{d}}

\newcommand{\x}{\ensuremath{\mathbf{x}}\xspace}

\newcommand{\xpath}{\overline{\x}}

\newcommand{\xs}{\ensuremath{\x_{s}}\xspace} 

\newcommand{\Norm}[1]{\left\lVert#1\right\rVert}

\newcommand{\xl}{\mathbf{x}_l} 

\newcommand{\tlv}{t_{lv}}
\newcommand{\tsv}{t_{sv}}

\newcommand{\xv}{\mathbf{x}_v}
\newcommand{\xbf}{\mathbf{x}}
\newcommand{\xa}{\mathbf{x}_a}
\newcommand{\xb}{\mathbf{x}_b}

\newcommand{\FK}{\textit{f-k}\xspace}
\newcommand{\fkmig}{\FK migration\xspace}

%% file: src0_abstract.tex
Time-of-Flight non-line-of-sight (ToF NLOS) imaging techniques provide state-of-the-art reconstructions of scenes hidden around corners by inverting the optical path of indirect photons scattered by visible surfaces and measured by picosecond resolution sensors. 
The emergence of a wide range of ToF NLOS imaging methods with heterogeneous formulae and hardware implementations obscures the assessment of both their theoretical and experimental aspects.
We present a comprehensive study of a representative set of ToF NLOS imaging methods by discussing their similarities and differences under common formulation and hardware. 
We first outline the problem statement under a common general forward model for ToF NLOS measurements, and the typical assumptions that yield tractable inverse models. 
We discuss the relationship of the resulting simplified forward and inverse models to a family of Radon transforms, and how migrating these to the frequency domain relates to recent phasor-based virtual line-of-sight imaging models for NLOS imaging that obey the constraints of conventional lens-based imaging systems. 
We then evaluate performance of the selected methods on hidden scenes captured under the same hardware setup and similar photon counts. 
Our experiments show that existing methods share similar limitations on spatial resolution, visibility, and sensitivity to noise when operating under equal hardware constraints, with particular differences that stem from method-specific parameters. 
We expect our methodology to become a reference in future research on ToF NLOS imaging to obtain objective comparisons of existing and new methods. 

%% file: src1_intro.tex
\section{Introduction}\label{sec:introduction}
Obtaining information of scenes that are not directly visible to an observer has been the subject of a wide variety of imaging techniques, commonly known as non-line-of-sight (NLOS) imaging \cite{Faccio2020non,jarabo2017recent,maeda2019recent}.
While differing in operational and computational aspects, NLOS imaging methods leverage a key physical principle---any visible surface is an omniscient observer of the indirect illumination from its surrounding objects, but which may be occluded from direct line of sight of the imaging device.
Capturing and analyzing the time of flight (ToF) of indirect illumination reflected by visible surfaces using ultra-fast imaging devices yields state-of-the-art 3D reconstructions of objects hidden around corners \cite{Liu2019phasor,OToole2018confocal,Lindell2019wave,Young2020Directional}. Such ToF NLOS imaging techniques have potential applications in medical imaging, autonomous driving, search and rescue operations, vision in adverse conditions, or unmanned exploration of inaccessible environments. 
The performance of NLOS imaging methods in real scenarios is, however, challenged by factors such as algorithmic assumptions, scene complexity, hardware limitations, and computational costs.  
These create operational and computational trade-offs, from which different ToF NLOS imaging methods have been devised. 

In this work we demonstrate that, while presented under heterogeneous formulae and method-motivated capture setups and scenarios, ToF NLOS imaging methods are built upon similar assumptions and algorithmic characteristics, and perform similarly when evaluated under equal hardware and scene conditions. 
For this purpose, we first formalize the underlying theoretical connections between a representative body of ToF NLOS imaging methods. We then  provide an experimental analysis of the main imaging trade-offs of these methods under a common set of simulated and real scenarios.

\paragraph{Scope of our contribution} In this work we provide a cross-sectional analysis, both theoretical and experimental, of a fundamental set of ToF NLOS imaging methods that build upon Radon transforms to define the NLOS acquisition process. 
While differing in capture and algorithmic aspects, we first demonstrate they all aim to invert different types of Radon transforms to image hidden scenes, and outline the theoretical parallelisms between these NLOS inverses and wave-based lens propagation in LOS imaging. 
Based on these theoretical grounds, we leverage the NLOS-LOS analogy to experimentally analyze performance aspects such surface visibility, resolution, and sensitivity to noise, and their relationship to scene and imaging parameters such as capture topology, photon counts, filtering properties, or geometry location.
We assemble qualitative and quantitative experimental cross-analyses of their performance in different conditions to support our theoretical study, showing that all methods generally display analogous imaging behavior, with differences that stem from algorithm-specific choices and capture limitations. 
We demonstrate performance changes and light transport behavior in ToF NLOS imaging pipelines are well explained and predicted by longstanding LOS optics principles, hoping to provide a more clear picture and foster novel methods on this direction. 

\paragraph{Relationship to previous analyses} Previous works \cite{Buttafava2015, OToole2018confocal, Liu2019analysis, Faccio2020non} provided valuable analyses of many aspects covered in our work under specific capture, algorithmic, and scene conditions, scratching the surface on the study of NLOS imaging performance. In our work we outline the theoretical relationship between a fundamental cross section of ToF NLOS methods and wave-based image formation principles, and cross-compare scene, algorithmic, and imaging configurations to reason about performance under the lens of LOS imaging principles. Conclusions of previous studies therefore match with a subset of the conclusions in this work, in which we provide a thorough and principled analysis of NLOS imaging performance from an optics-based perspective, under the same hardware and scene conditions. 

%% file: src2_related_work.tex
\section{Related work}
A main operational distinction on the body of NLOS imaging works is between steady-state and time-resolved methods. {Steady-state} methods typically use passive illumination, and capture indirect illumination at visible surfaces under large exposure times compared to the hidden scene size. They ignore any temporal information of light propagation, and rely on analyzing spatial gradients of indirect illumination produced by the hidden scene. This turns hidden scene reconstruction into a very ill-posed problem that demands strong assumptions. Early methods dated from 1982 \cite{cohen1982anti} and the more recent work by Torralba and Freeman \cite{torralba2012accidental} exploit pin-shaped occlusions to visualize hidden scenes, requiring very specific scene layouts. Supporting more general scenes usually limits the applicability to classification or tracking tasks \cite{Klein2016SR}, requires operating in lower-dimensional imaging spaces \cite{bouman2017turning,saunders2019computational}, or demands strong regularization on the imaging process \cite{seidel2019corner}. Recent work by Chen et al. \cite{Chen2019steady} provided actual images of the hidden scene using conventional cameras along with \textit{active} illumination.

In contrast to steady-state methods, {time-resolved} NLOS methods reconstruct the hidden scene by analyzing indirect light at ultra-fast temporal resolutions---from nano- to femto-seconds---by means of active illumination. 
Working at ultra-fast temporal resolutions develops a dense amount of time-resolved information of light propagation that in steady-state methods is integrated (and lost) over a single image. This transient indirect transport is directly related to geometric, material, and spatial information of the hidden scene. Early methods used expensive streak cameras \cite{Velten2012nc} at femtosecond resolution to reconstruct meter-scale geometry, while recent works used interferometry to obtain similar temporal resolutions to recover hidden geometry at micrometer scales \cite{Xin2019theory}. Correlation-based imaging has been proven an effective and inexpensive technique for room-sized NLOS reconstructions \cite{Heide2014diffuse,kadambi2016occluded}. A few works proposed to use deep learning \cite{GrauCVPR2020} or transient rendering \cite{iseringhausen2020non} to estimate the hidden scene, or even tracking objects occluded after two corners \cite{Yi2021SIGA} by using differentiable time-resolved transport. 

Within the vast amount of time-resolved approaches, combining single-photon avalanche diodes (SPADs) and pulsed laser illumination has fostered a large body of \textit{time-of-flight} (ToF) NLOS imaging methods \cite{Buttafava2015,tsai2017geometry, LaManna2018error,OToole2018confocal,Lindell2019wave,Xin2019theory,ahn2019convolutional,Liu2019phasor} that capture and analyze indirect light at picosecond resolution. 
Recent advances on ToF methods have demonstrated unprecedented results on reconstructing and analyzing complex meter-scale NLOS scenes. 
Their ability to deliver sharp 3D reconstructions of hidden cluttered configurations has placed them as a promising and robust pathway for applications in real-world scenarios. 
In their most typical form, these methods actively illuminate a visible diffuse \emph{relay} surface with an ultra-short laser pulse, and analyze the time of flight of indirect photons of a hidden scene at a meter-scale area of the relay surface. 
ToF NLOS imaging methods computationally exploit this developed transport to demultiplex rich information of the hidden scene, such as geometric density \cite{Buttafava2015,OToole2018confocal,Lindell2019wave}, surface materials and normals \cite{Xin2019theory,Young2020Directional}, transient illumination \cite{Liu2019phasor}, or light transport components \cite{Marco2021NLOSvLTM}. Recent works unveiled parallelisms between ToF NLOS imaging and traditional LOS wave optics  \cite{Liu2019phasor,Liu2020phasor}, enabling interactive \cite{nam2021low} and real-time operation times \cite{liao2021fpga} by means of fast optics-based propagators. These parallelisms stem from mathematical equivalences between longstanding optics principles that define lens-based LOS image formation, and backprojection solutions to different inverse Radon transforms that govern most ToF NLOS imaging methods. While implicit in a wide number of ToF methods, a thorough cross-sectional analysis of these NLOS-LOS parallelisms and their potential on predicting NLOS imaging performance remains obscured.

%% file: src3_problem_statement.tex
\section{Problem statement}
\label{sec:problem_statement}
\input{tab_notation}
\begin{figure*}
    \centering
	{\def\svgwidth{\textwidth}
	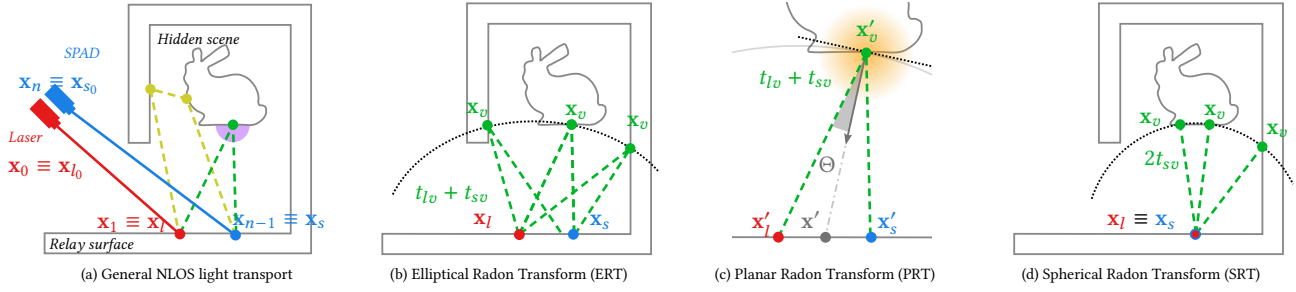}
    \caption{
    (a) ToF NLOS imaging setups measure multiply scattered photons following paths between a laser device at $\xbf_0$, a laser target $\xbf_1$, sensor target $\xbf_{n-1}$, and sensor device at $\xbf_n$. Classic approaches assume perfect calibration of laser and sensor devices and targets (red and blue), third-bounce-only occlusion-free transport (green), and diffuse surface reflectance (purple). 
    Under these assumptions, time-resolved NLOS measurements can be modeled by different types of Radon transforms. 
    (b) NLOS light transport of photons with time of flight $t_{lv}+t_{sv}$ under unconstrained $\xl$ and $\xs$ is modeled by the elliptical Radon transform (ERT), whose inverse is typically approximated via filtered backprojection (\cref{sec:ERT}). 
    (c) By assuming planar locality at surface points $\xv$, NLOS light transport can be approximated by the planar Radon transform (PRT), whose inverse can be solved analytically by means of Laplacian-filtered backprojection (\cref{sec:PRT}).
    (d) Confocal setups co-locate $\xl \equiv \xs$, under which the ERT becomes a spherical Radon transform (SRT); existing methods leveraged this constraint to implement efficient solvers for the inverse SRT (\cref{sec:PRT}). 
    }
    \label{fig:nlos_transport_inverses}
\end{figure*}
In the following we outline the light transport model that defines time-resolved illumination captured on NLOS setups, and summarize the main assumptions made by NLOS reconstruction methods over such model to provide tractable reconstruction algorithms. 
In \cref{sec:primal_NLOS_models} we provide an overview of the different forward transport models that result from these assumptions, and the corresponding inverses proposed by existing NLOS imaging methods. 
In \cref{sec:frequency_NLOS_models} we summarize the relationship between these inverses and wave-based propagation operators of traditional line-of-sight imaging models. \cref{tab:notation} contains a glossary of the most relevant terms and symbols used throughout the article. 

\subsection{Transient light transport in NLOS setups} 
\paragraph{Transient path integral}
{ToF} NLOS imaging methods capture time-resolved indirect illumination produced by a hidden scene on a visible relay surface after illuminating and measuring a set of discrete points on such surface with an ultra-fast imaging device. Any time-resolved measurement $H$ on the relay surface is defined by the transient path integral formulation \cite{Jarabo2014} as
\begin{gather}
    H = \int\limits_{\Psi} \int\limits_{\mathcal{T}} L(\x_0, \Delta t_0) \mathfrak{T}(\xpath, \overline{\Delta \mathbf{t}}) W(\x_n, \Delta t_n) \diff \mu(\overline{\Delta \mathbf{t}}) \diff \mu(\xpath), 
    \label{eq:path_integral}
\end{gather}
where 
$\xpath = \langle \x_0 \dots \x_n\rangle$ represents a $(n+1)-$vertex path between an illumination device at $\x_0$ and a sensor device at $\x_n$ (\cref{fig:nlos_transport_inverses}a); 
$\Psi$ is the space of all possible light paths $\xpath$;
$\mathcal{T}$ is the space of temporal delays $\overline{\Delta \x}$ of such light paths;
$L$ represents the emitter intensity; 
$W$ represents sensor sensitivity to a path with a time of flight $\Delta t_n$;
and $\mu(\xpath)$ denotes integration at path vertices.
$\mathfrak{T}(\xpath, \overline{\Delta \mathbf{t}})$ represents the throughput of path $\xpath$ as a result of visibility, geometric attenuation, and surface reflectance between all path vertices, having
\begin{align}
    \label{eq:path_throughput}
    \mathfrak{T}(\xpath, \overline{\Delta \mathbf{t}}) &= f(\xpath) G(\xpath) T(\xpath), \\
\label{eq:reflectance_prod}
f(\xpath) &= \prod\limits_{j=1}^{n-1} f(\x_{j-1}, \x_{j}, \x_{j+1}), \\
\label{eq:geometry_prod}
G(\xpath) &= \prod\limits_{j=0}^{n-1} G(\x_{j},\x_{j+1}), \\
\label{eq:visibility_prod}
T(\xpath) &= \prod\limits_{j=0}^{n-1} T(\x_{j}, \x_{j+1}),
\end{align}
where $f(\x_{j-1}, \x_j, \x_{j+1})$ models surface reflectance at $\x_j$, and $G(\x_j, \x_{j+1})$ and $T(\x_j, \x_{j+1})$ model geometric attenuation and binary visibility between two surface points $\x_j, \x_{j+1}$, respectively. 
\paragraph{NLOS impulse response function} 
Typical ToF NLOS imaging devices consist of an ultra-fast laser source at $\x_0$ that emits a beam of light targeted at specific points $\x_1$ on the relay surface, and a single-photon avalanche-diode (SPAD) sensor at $\x_n$ measuring the returning signal at a small region centered at $\x_{n-1}$ on the relay surface. 
Laser and sensor devices remain static during the entire capture process, therefore $\x_0, \x_n$ are constant for all measurements. 

Under spatio-temporal delta emission $L(\x_1, \Delta t_0) \equiv \delta(\xbf-\x_1, \Delta t_0 - t)$ and a spatio-temporal delta response function for the sensor $W(\xbf_n, \Delta t_n) \equiv \delta(\xbf-\x_{n-1}, \Delta t_n - t)$, 
delta emission and sensitivity $L \equiv \delta(\xbf-\x_1, \Delta t_0 - t), W(\xbf_n, \Delta t_n) \equiv \delta(\xbf-\x_{n-1}, \Delta t_n - t)$ in both spatial and the temporal domain, we can rewrite \cref{eq:path_integral,eq:path_throughput} as the following \emph{continuous} forward model
\footnote{Throughout the article we distinguish between continuous and matricial form of some equations through subscripts $c$ and $m$ in the equation numbers; this helps emphasize the relation between both forms. When no subscript is used, we are referring to both versions of the same equation.}
\begin{subequations}
\label{eq:general_formation}
\begin{gather}
    \label{eq:general_formation_cont}
    \tag{\theequation\textsubscript{c}} 
    H(\x_1, \x_{n-1}, t) = \int\limits_{\Psi}  {f}(\xpath) {G}(\xpath) {T}(\xpath) \delta(t-\Delta t_n) \diff \mu(\xpath).
\end{gather}
\end{subequations}
where $\Delta t_k = \sum_{j=1}^n \Norm{\x_j-\x_{j-1}}/\sol$ is the time of flight of path $\xpath$, with $\sol$ the speed of light. 

%
Throughout this article we often rely on the matricial form of continuous equations for algorithmic analysis. We can express \cref{eq:general_formation_cont} in  matricial form as
\begin{align}
    \label{eq:general_formation_mat}
    \tag{\theequation\textsubscript{m}} 
    \mat{h} & = \mat{A}\mat{T}\mat{G}\mat{f}, 
\end{align}
where $\mat{h}$, $\mat{A}$, $\mat{T}$, $\mat{G}$, and $\mat{f}$ represent discretized versions of $H$, $\delta$ $T$, $G$, and $f$, respectively, for a set of measurements $f(\xpath)$ with known $\x_0, \x_1, \x_{n-1}$ and $\x_n$. $H\equiv\mat{h}$ is known as the \textbf{impulse response function} of the scene.

Inferring the scene geometry from a measured $H\equiv \mat{h}$ is an extremely ill-posed problem, as all terms in \cref{eq:general_formation} depend on a potentially infinite number of paths $\xpath \in \Psi$ with unknown subpaths $\x_2...\x_{n-2}$.
In the following we summarize well-established assumptions made by NLOS imaging methods to provide tractable reconstruction algorithms based on simplified transport models.

\subsection{Light transport assumptions in NLOS imaging}
To provide tractable reconstructions of hidden scenes from a measured impulse response function $H$, NLOS imaging methods typically make several assumptions about forward light transport (\cref{eq:general_formation}), specifically:
\begin{enumerate}[label=\textit{Asm.\arabic*}]
\item \label{asm:direct} Three-bounce transport: There are no indirect interreflections between hidden surfaces, constraining light transport to three-bounce paths $\xpath~=~\langle\x_0,\x_1,\x_2,\x_3,\x_4\rangle$  (\cref{fig:nlos_transport_inverses}a, green).
\item \label{asm:lambertian} Lambertian scattering: Surfaces scatter light equally in all directions, therefore $f(\x_{j-1}, \x_{j}, \x_{j+1})$ (\cref{eq:reflectance_prod}) loses dependency on neighboring points, and $f(\x_j)$ represents the albedo for all $\x_j \in \xpath$ on a surface. 
\item \label{asm:occlusions} No occlusions: All surface points in the hidden scene are visible from both the laser and sensor locations on the relay surface. This removes the $T \equiv \mat{T}$ dependency of \cref{eq:general_formation}.
\item \label{asm:attenuation} Slow-varying attenuation: 
It is common to assume the geometry term $G(\x_{j},\x_{j+1})$ (\cref{eq:geometry_prod}) varies rather slowly through the hidden scene, which allows to pull a constant term $\overline{G}$ out of \cref{eq:general_formation_cont}. We specify what methods relax this assumption when needed.  
\item \label{asm:calibration} Calibration: Both imaging device and relay surface are calibrated in the capture process. Therefore, $\x_0, \x_1, \x_3$, and $\x_4$ are known, and their $f$ values are either known or assumed constant. 
\end{enumerate}
Under calibrated or constant relay surface reflectance $f(\x_1)$ and $f(\x_3)$, solving \cref{eq:general_formation} for $f(\xpath)$ boils down to determining $f(\x_2)$. 

\ref{asm:direct} to \ref{asm:calibration} define \textbf{the problem tackled by ToF NLOS imaging methods: estimating hidden surface albedos $f(\x_2)$ by inverting the light transport models that result from applying such assumptions to general light transport in NLOS scenes (\cref{eq:general_formation}}).

In \cref{sec:primal_NLOS_models} we show that ToF NLOS imaging methods stem from similar light transport models that result from these assumptions, while they differ in 1) model simplifications and re-formulations based on the capture topology; and 2) the filtering procedures used by each method. 
We show that depending on the capture topology, the resulting transport models correspond to different types of Radon transforms, to which NLOS imaging methods provide inverse solutions with trade-offs between computational performance, generality, and reconstruction quality. 
In \Cref{sec:frequency_NLOS_models} we summarize the narrow relationship of the frequency-domain counterparts of these light propagation models to wave propagation integrals \cite{Liu2019analysis,Lindell2019wave}. 
Based on these theoretical connections, in \cref{sec:evaluation,sec:eval_capture} we perform an experimental cross-analysis of NLOS imaging methods in simulated and captured scenarios. 

To improve readability, we use letter subindices for the vertices of three-bounce paths in NLOS scenes, having $\xpath = \langle\x_{l_0}, \xl, \xv, \xs, \x_{s_0} \rangle$, where $\x_{l_0}$ and $\x_{s_0}$ are the location of the laser and SPAD devices, $\xl$ and $\xs$ are laser and SPAD positions at the relay surface, and $\xv$ are hidden scene points. 
As temporal delays between the device and the relay surface $\x_{l_0}\rightarrow \xl, \xs\rightarrow \x_{s_0}$ are constant throughout the capture process, impulse response functions are typically time-normalized at the relay surface, so that such temporal delays are not accounted for in $H(\xl, \xs, t)$. We adopt this convention throughout the article without loss of generality.

%% file: tab_notation.tex
\begin{table}[t]\small 
    \centering
    \caption{Notation used throughout the paper. Note in several cases continuous functions are defined along their respective matricial equivalent such as $G(\cdot) \equiv \mat{G}$.}
    \begin{tabularx}{\columnwidth}{p{2cm} X}    
	\toprule  
	$t$                                                     & Time. \\
	$\freq$                                                 & Frequency. \\
	$c$                                                     & Speed of light. \\
	$V$ 											        & Hidden scene volume. \\
	$L,S$ 								                    & Illumination and sensor relay surfaces.\\	
	$\xa, \xb, \xv\! \in\! V$								& Points in the hidden scene. \\
	$\xl\!\in\!L, \xs\!\in\!S$					            & Points at visible planes $L$ and $S$.\\
	\midrule
        $H(\cdot) \equiv \mat{h}$                                   & Impulse response of the hidden scene. \\
	$f(\cdot) \equiv \mat{f}$                                 & Reflectance at a point or a light path. \\
	$T(\cdot) \equiv \mat{T}$                               & Binary visibility function. \\
	$G(\cdot) \equiv \mat{G}$                               & Geometric attenuation function. \\
	$K(\cdot) \equiv \mat{K}$                               & Filter over a set of dimensions. \\
	$\delta(\cdot) \equiv \mat{D}$                                         & Dirac delta function. \\
	$\mat{A}_\square, \mat{A}^\transpose_\square$           & Forward and backprojection operators. \\
    $\Fspace{P}_l(\cdot), \Fspace{P}_s(\cdot)$                                              & Complex-valued phasors at an illumination $\xl$ or sensor location $\xs$.\\
    ${\mathrm{R}_t{\cdot} \equiv \mat{R}_t}$, 
    ${\mathrm{R}_z{\cdot} \equiv \mat{R}_z}$          & Reparameterization functions for time $t$ and depth $z$ domains. \\
    $\mathcal{R}_{A \rightarrow B}(\cdot)$                  & RSD propagation integral. \\
	\bottomrule
    \end{tabularx}
    \label{tab:notation}
\end{table}

%% file: figures/nlos_transport_inverses_svg-tex.pdf_tex
\begingroup%
  \makeatletter%
  \providecommand\color[2][]{%
    \errmessage{(Inkscape) Color is used for the text in Inkscape, but the package 'color.sty' is not loaded}%
    \renewcommand\color[2][]{}%
  }%
  \providecommand\transparent[1]{%
    \errmessage{(Inkscape) Transparency is used (non-zero) for the text in Inkscape, but the package 'transparent.sty' is not loaded}%
    \renewcommand\transparent[1]{}%
  }%
  \providecommand\rotatebox[2]{#2}%
  \newcommand*\fsize{\dimexpr\f@size pt\relax}%
  \newcommand*\lineheight[1]{\fontsize{\fsize}{#1\fsize}\selectfont}%
  \ifx\svgwidth\undefined%
    \setlength{\unitlength}{518.58891678bp}%
    \ifx\svgscale\undefined%
      \relax%
    \else%
      \setlength{\unitlength}{\unitlength * \real{\svgscale}}%
    \fi%
  \else%
    \setlength{\unitlength}{\svgwidth}%
  \fi%
  \global\let\svgwidth\undefined%
  \global\let\svgscale\undefined%
  \makeatother%
  \begin{picture}(1,0.21625011)%
    \lineheight{1}%
    \setlength\tabcolsep{0pt}%
    \put(0,0){\includegraphics[width=\unitlength,page=1]{nlos_transport_inverses_svg-tex.pdf}}%
    \put(0.66499486,0.04973708){\color[rgb]{0.10196078,0.50196078,0.89803922}\makebox(0,0)[lt]{\lineheight{1.25}\smash{\begin{tabular}[t]{l}$\xs'$\end{tabular}}}}%
    \put(0.58654827,0.04960273){\color[rgb]{0.89803922,0.10196078,0.10196078}\makebox(0,0)[rt]{\lineheight{1.25}\smash{\begin{tabular}[t]{r}$\xl'$\end{tabular}}}}%
    \put(0.64916296,0.19195694){\color[rgb]{0,0.70980392,0.15294118}\makebox(0,0)[lt]{\lineheight{1.25}\smash{\begin{tabular}[t]{l}$\xv'$\end{tabular}}}}%
    \put(0.63164511,0.15692047){\color[rgb]{0,0.70980392,0.15294118}\makebox(0,0)[rt]{\lineheight{1.25}\smash{\begin{tabular}[t]{r}$\tlv + \tsv$\end{tabular}}}}%
    \put(0,0){\includegraphics[width=\unitlength,page=2]{nlos_transport_inverses_svg-tex.pdf}}%
    \put(0.62155548,0.09146754){\color[rgb]{0.45098039,0.45098039,0.45098039}\makebox(0,0)[lt]{\lineheight{1.25}\smash{\begin{tabular}[t]{l}$\Theta$\end{tabular}}}}%
    \put(0.62316787,0.04978355){\color[rgb]{0.45098039,0.45098039,0.45098039}\makebox(0,0)[rt]{\lineheight{1.25}\smash{\begin{tabular}[t]{r}$\x'$\end{tabular}}}}%
    \put(0,0){\includegraphics[width=\unitlength,page=3]{nlos_transport_inverses_svg-tex.pdf}}%
    \put(0.4595597,0.05198137){\color[rgb]{0.10196078,0.50196078,0.89803922}\makebox(0,0)[t]{\lineheight{1.25}\smash{\begin{tabular}[t]{c}$\xs$\end{tabular}}}}%
    \put(0,0){\includegraphics[width=\unitlength,page=4]{nlos_transport_inverses_svg-tex.pdf}}%
    \put(0.39169073,0.01208818){\color[rgb]{0,0,0}\makebox(0,0)[t]{\lineheight{1.25}\smash{\begin{tabular}[t]{c}\scriptsize{(b) Elliptical Radon Transform (ERT)}\end{tabular}}}}%
    \put(0,0){\includegraphics[width=\unitlength,page=5]{nlos_transport_inverses_svg-tex.pdf}}%
    \put(0.37674762,0.07150973){\color[rgb]{0,0.70980392,0.15294118}\makebox(0,0)[rt]{\lineheight{1.25}\smash{\begin{tabular}[t]{r}$\tlv + \tsv$\end{tabular}}}}%
    \put(0.36038834,0.13496569){\color[rgb]{0,0.70980392,0.15294118}\makebox(0,0)[lt]{\lineheight{1.25}\smash{\begin{tabular}[t]{l}$\xv$\end{tabular}}}}%
    \put(0.433355,0.13332861){\color[rgb]{0,0.70980392,0.15294118}\makebox(0,0)[lt]{\lineheight{1.25}\smash{\begin{tabular}[t]{l}$\xv$\end{tabular}}}}%
    \put(0.47919317,0.1258448){\color[rgb]{0,0.70980392,0.15294118}\makebox(0,0)[lt]{\lineheight{1.25}\smash{\begin{tabular}[t]{l}$\xv$\end{tabular}}}}%
    \put(0.62602256,0.01208818){\color[rgb]{0,0,0}\makebox(0,0)[t]{\lineheight{1.25}\smash{\begin{tabular}[t]{c}\scriptsize{(c) Planar Radon Transform (PRT)}\end{tabular}}}}%
    \put(0.86081332,0.05197655){\color[rgb]{0,0,0}\makebox(0,0)[t]{\lineheight{1.25}\smash{\begin{tabular}[t]{c}$\textcolor[HTML]{E51A1A}{\xl} \equiv \textcolor[HTML]{1A80E5}{\xs}$\end{tabular}}}}%
    \put(0.86023279,0.01208336){\color[rgb]{0,0,0}\makebox(0,0)[t]{\lineheight{1.25}\smash{\begin{tabular}[t]{c}\scriptsize{(d) Spherical Radon Transform (SRT)}\end{tabular}}}}%
    \put(0,0){\includegraphics[width=\unitlength,page=6]{nlos_transport_inverses_svg-tex.pdf}}%
    \put(0.88926584,0.09654203){\color[rgb]{0,0.70980392,0.15294118}\makebox(0,0)[rt]{\lineheight{1.25}\smash{\begin{tabular}[t]{r}$2\tsv$\end{tabular}}}}%
    \put(0.85916075,0.12781391){\color[rgb]{0,0.70980392,0.15294118}\makebox(0,0)[lt]{\lineheight{1.25}\smash{\begin{tabular}[t]{l}$\xv$\end{tabular}}}}%
    \put(0.90696359,0.13332861){\color[rgb]{0,0.70980392,0.15294118}\makebox(0,0)[lt]{\lineheight{1.25}\smash{\begin{tabular}[t]{l}$\xv$\end{tabular}}}}%
    \put(0.95317624,0.11985121){\color[rgb]{0,0.70980392,0.15294118}\makebox(0,0)[lt]{\lineheight{1.25}\smash{\begin{tabular}[t]{l}$\xv$\end{tabular}}}}%
    \put(0.37276483,0.05289243){\color[rgb]{0.89803922,0.10196078,0.10196078}\makebox(0,0)[t]{\lineheight{1.25}\smash{\begin{tabular}[t]{c}$\xl$\end{tabular}}}}%
    \put(0,0){\includegraphics[width=\unitlength,page=7]{nlos_transport_inverses_svg-tex.pdf}}%
    \put(0.08834938,0.17581257){\color[rgb]{0.10196078,0.50196078,0.89803922}\makebox(0,0)[rt]{\lineheight{1.25}\smash{\begin{tabular}[t]{r}\emph{\scriptsize{SPAD}}\\$\xbf_n \equiv \x_{s_0}$\end{tabular}}}}%
    \put(0.18690722,0.05236914){\color[rgb]{0.10196078,0.50196078,0.89803922}\makebox(0,0)[lt]{\lineheight{1.25}\smash{\begin{tabular}[t]{l}$\xbf_{n-1} \equiv \xs$\end{tabular}}}}%
    \put(0,0){\includegraphics[width=\unitlength,page=8]{nlos_transport_inverses_svg-tex.pdf}}%
    \put(0.1381741,0.04986091){\color[rgb]{0.89803922,0.10196078,0.10196078}\makebox(0,0)[rt]{\lineheight{1.25}\smash{\begin{tabular}[t]{r}$\xbf_1 \equiv \xl$\end{tabular}}}}%
    \put(0.02045382,0.11334007){\color[rgb]{0.89803922,0.10196078,0.10196078}\makebox(0,0)[lt]{\lineheight{1.25}\smash{\begin{tabular}[t]{l}\emph{\scriptsize{Laser}}\\$\xbf_0 \equiv \x_{l_0}$\end{tabular}}}}%
    \put(0.05097273,0.03459526){\color[rgb]{0,0,0}\makebox(0,0)[lt]{\lineheight{1.25}\smash{\begin{tabular}[t]{l}\emph{\scriptsize{Relay surface}}\end{tabular}}}}%
    \put(0,0){\includegraphics[width=\unitlength,page=9]{nlos_transport_inverses_svg-tex.pdf}}%
    \put(0.13095251,0.18651456){\color[rgb]{0,0,0}\makebox(0,0)[lt]{\lineheight{1.25}\smash{\begin{tabular}[t]{l}\emph{\scriptsize{Hidden scene}}\end{tabular}}}}%
    \put(0,0){\includegraphics[width=\unitlength,page=10]{nlos_transport_inverses_svg-tex.pdf}}%
    \put(0.15306221,0.01208818){\color[rgb]{0,0,0}\makebox(0,0)[t]{\lineheight{1.25}\smash{\begin{tabular}[t]{c}\scriptsize{(a) General NLOS light transport}\end{tabular}}}}%
    \put(0,0){\includegraphics[width=\unitlength,page=11]{nlos_transport_inverses_svg-tex.pdf}}%
  \end{picture}%
\endgroup%

%% file: src4_NLOS_inverses.tex
\section{Inverting NLOS transient light transport}
\label{sec:primal_NLOS_models}
Here we show that the light transport models that result from assumptions \ref{asm:direct} to \ref{asm:calibration} made on the general light transport \cref{eq:general_formation} correspond to different types of Radon transforms; such transforms have been widely studied in other applications \cite{moon2014determination,deans2007radon,Xu2005universal} and motivate the different inverse approximations introduced in the NLOS imaging literature.
Analogously to \cref{eq:general_formation}, we summarize and relate these Radon transforms in both continuous and matricial form, showing how different methods describe ways to invert them to image the hidden scene.

\subsection{Elliptical Radon Transform} 
\label{sec:ERT}
Under \ref{asm:direct} to \ref{asm:calibration} and unconstrained configurations of illumination and sensor points $\xl, \xs$ (\cref{fig:nlos_transport_inverses}b), the continuous light transport model (\cref{eq:general_formation_cont}) is equivalent to the following expression
\begin{subequations}
\label{eq:ERT}
\begin{align}
\tag{\theequation\textsubscript{c}}\label{eq:ERT_cont}
    H(\xl,\xs,t) & = \overline{G} \int\delta\left(t-(\tlv + \tsv)\right) f(\xv) \diff\xv.
\end{align}
\end{subequations}
The Dirac delta term $\delta(\cdot)$ constrains propagation to three-bounce light paths with time-of-flight $\tlv+\tsv$ with $t_{ab} = \Norm{\xa-\xb}/\sol$. Note due to time-normalization of $H$ at the relay surface, time of flight of relay-device paths $t_{l_0v} = \Norm{\x_{l_0}-\xv}/c$ and $t_{s_0v} = \Norm{\x_{s_0}-\xv}/c$ are not accounted for.
Under slow-varying attenuation (\ref{asm:attenuation}), ${\overline{G} = \int G(\xl,\xs,\xv) \diff \xv}$ is a constant term approximating the average attenuation between all $\xv$ points in the scene, and specific points $\xl$ and $\xs$ at the relay surface. This forward model corresponds to an \textit{elliptical 3D Radon transform} (ERT) \cite{moon2014determination} (\cref{fig:nlos_transport_inverses}b).

In matricial form, \cref{eq:ERT_cont} can be expressed as
\begin{align}
\tag{\theequation\textsubscript{m}} \label{eq:ERT_mat}
    \mat{h} &= \overline{G} \mat{A}_\mathrm{ert} \mat{f},
\end{align}
where $\mathbf{A}_\mathrm{ert}$ is the ERT forward operator that relates every location $\xv$ in $\mathbf{f}$ to the third-bounce light paths with $\xl$ and $\xl$ illumination and sensor coordinates in $\mathbf{h}$. 

The general ERT has no known analytical inverse solution for $f$. Velten et al. \cite{Velten2012nc} proposed a filtered backprojection solution where $f(\xv)$ is computed as the contribution of all $H$ values with the time of flight of third-bounce paths, having 
\begin{subequations}
\label{eq:inv_ERT}
\begin{align}
\tag{\theequation\textsubscript{c}} 
\label{eq:inv_ERT_cont}
    f(\xv)\! &\approx\! K(\xv) \conv \!\! \iint\limits_{S\,L} H(\xl,\xs,\tlv+\tsv) \diff \xl \diff \xs, \\
\tag{\theequation\textsubscript{m}}
\label{eq:inv_ERT_mat} 
    \mat{f} &\approx \mat{K}_{\xv} \mat{A}_\mathrm{ert}^\transpose \mat{h}, 
\end{align}
\end{subequations}
where $K(\xv)\equiv \mat{K}_{\xv}$ is a spatial filter over of scene points $\xv$. Values of $H$ with any time of flight $t'$ contribute to an ellipsoidal manifold in the reconstruction space defined by all points at a total distance $ct'$ from foci $\xl, \xs$ (\cref{fig:nlos_transport_inverses}b). 
In matricial form (\cref{eq:inv_ERT_mat}), $\mat{A}_\mathrm{ert}^\transpose$ is known as the \textit{backprojection operator} (i.e. the transpose of the forward operator $\mat{A}_\mathrm{ert}$).
Typical choices for $K(\xv)\equiv \mat{K}_{\xv}$ in most backprojection implementations are Laplacian or Laplacian-of-Gaussian filters, which can be derived as analytical solutions of the planar Radon transform.
%

\subsection{Planar Radon Transform}
\label{sec:PRT}
One approach to obtain an approximate \textit{analytical} inverse of \cref{eq:ERT} for $f$ is to locally approximate the ellipses by planes, as illustrated in \cref{fig:nlos_transport_inverses}c. By shifting the coordinate system so that the origin is at $\xv'=(x',y',z')$, the impulse response function is well approximated by a \emph{planar 3D Radon transform} \cite{Liu2019analysis},
\begin{subequations}
\label{eq:PRT}
\begin{align}
\tag{\theequation\textsubscript{c}} 
\label{eq:PRT_cont}
\begin{aligned}
H(\xl',\xs',t-(\tlv+\tsv))\approx \\
    \approx \iiint{f(\x_v')} 
    \delta(k_x x'+ k_y y'+ k_z z') 
    dx'dy'dz',
\end{aligned}
\end{align}
\begin{align}
\tag{\theequation\textsubscript{m}} 
\label{eq:PRT_mat}
    \mat{h} = \mat{A}_\mathrm{prt} \mat{f},
\end{align}
\end{subequations}
where the plane orientation $(\Theta,\Phi)$ is defined such that its normal vector points to the center of the ellipsoid $\xbf' = (\xl'+\xs')/2$, and 
\[ k_x = \sin\Theta \cos\Phi, k_y = \sin\Theta\sin\Phi, k_z = \cos\Theta.\] 
Note \cref{fig:nlos_transport_inverses}c only displays $\Theta$ as it is a 2D schematic.
In matricial form (\cref{eq:PRT_mat}), $\mat{A}_\mathrm{prt}$ denotes the forward operator for the PRT. 

The planar Radon transform can be solved analytically by a \textit{time-filtered} backprojection as
\begin{subequations}
\label{eq:inv_PRT}
\begin{align}
\tag{\theequation\textsubscript{c}} 
\label{eq:inv_PRT_cont}
    f(\xv) &\approx\!\!\iint\limits_{S\,L}\nabla_t^2 H(\xl,\xs,\tlv+\tsv) \diff \xl \diff \xs,\\
\tag{\theequation\textsubscript{m}} 
\label{eq:inv_PRT_mat}
    \mat{f} &\approx \mat{A}_\mathrm{prt}^\transpose \mat{K}_{\nabla,t} \mat{h},
\end{align}
\end{subequations}
where $\nabla_t^2$ is a Laplacian operator in the temporal domain of the impulse response function, \emph{before} the backprojection integral. This is reflected in matricial form (\cref{eq:PRT_mat}) by applying the operator $\mat{K}_{\nabla,t}$ first, then a backprojection $\mat{A}_\mathrm{prt}^\transpose$ over the time-filtered data. 
The analytical solution of the inverse PRT motivates the choice of Laplacian and Laplacian-of-Gaussian filters for $K(\xv)$ when solving \cref{eq:inv_ERT}. We analyze their impact in reconstruction performance in \cref{sec:evaluation,sec:eval_capture}. 

\subsection{Spherical Radon Transform} 
\label{sec:SRT}
\cref{eq:ERT,eq:PRT} describe forward NLOS transport with no constraints in the location of lasers and sensors. In a confocal setup where $\xl\equiv\xs$,  \cref{eq:ERT} turns into \cite{OToole2018confocal} (\cref{fig:nlos_transport_inverses}d)
\begin{align}
    \label{eq:SRT}
    H(\xs, t)= \widetilde{G} \int \delta(2\tsv-t) f(\xv) \diff\xv,
\end{align}
which corresponds to a \textit{spherical Radon transform} (SRT) \cite{Xu2005universal,Tasinkevych2014circular}. 
O'Toole et al. \cite{OToole2018confocal} showed that, in this case, the $G$ term in \cref{eq:general_formation} can be pulled out as a scale factor $G \equiv \widetilde{G} = (c\tsv)^4$ without the need for the approximation introduced by \ref{asm:attenuation}.  
More importantly, the SRT (\cref{eq:SRT}) can be re-parameterized to express it as a 3D shift-invariant convolution, denominated the Light-Cone Transform (LCT), as
\begin{subequations}
\label{eq:LCT}
\begin{align}
\tag{\theequation\textsubscript{c}} 
\label{eq:LCT_cont}
\reparam{R}_t\{H\}(\bar\xbf_s, \bar t) & = \int a(\bar\xbf_v-\bar\xbf_s) \reparam{R}_z\{f\}(\bar\xbf_v)  \diff \bar\xbf_v,\\
\begin{split}
\label{eq:LCT_mat}
\mat{R}_t \mat{h}   &= \mat{A}_\mathrm{lct} \mat{R}_z \mat{f} = \mat{a}_\mathrm{lct} \ast (\mat{R}_z \mat{f}),
\end{split}
\tag{\theequation\textsubscript{m}} 
\end{align}
\end{subequations}
where $\bar\xbf_v$ and $\bar\xbf_s$ correspond to the reparameterized spaces of $\xv$ and $\langle\xs,t \rangle$; $\reparam{R}_t\{\cdot\} \equiv \mat{R}_t$ and $\reparam{R}_z\{\cdot\} \equiv \mat{R}_z$ are re-parameterizations of the time and depth dimensions, respectively;  $a \equiv \mat{A}_\mathrm{lct}$ constrains light transport to three-bounce paths over the re-parameterized $\bar\xbf_v$ and $\bar\xbf_s$. Light transport constrained by $a \equiv \mat{A}_\mathrm{lct}$ can be computed as a shift-invariant 3D convolution by kernel $\mat{a}_\mathrm{lct}$ over the reparameterized space of $\bar\xbf_v$. Please refer to \cite{OToole2018confocal} for a more detailed description of the different terms.

The hidden scene $\mat{f}$ can be efficiently estimated by inverting \cref{eq:LCT} via deconvolutions in the Fourier domain \cite{OToole2018confocal}, as
\begin{subequations}
\label{eq:inv_LCT}
\begin{align}
\tag{\theequation\textsubscript{c}} 
\label{eq:inv_LCT_cont}
    \Fspace{f}(\freq_{\xv}) &\approx \reparam{R}_{z}^{-1} \Bigl\{\Fspace{a}^{-1}(\freq_{\bar\xbf_v-\bar\xbf_s}) \reparam{R}_t\{\Fspace{H}\}(\freq_{\bar\xbf_s})\Bigr\}\left(\freq_{\xv}\right),\\
\tag{\theequation\textsubscript{m}} 
\label{eq:inv_LCT_mat}
    \mat{f} &\approx \minv{\mat{R}}_z \minv{\mat{F}} \widehat{\mat{A}}_{\mathrm{ilct}} \mat{F} \mat{R}_t,\mat{h},
\end{align}
\end{subequations}
where $\mat{F}$ represents the discrete Fourier transform in matricial form, and $\widehat{\mat{A}}_{\mathrm{ilct}}$ is a deconvolution based on Wiener filtering
\begin{align}
\label{eq:LCT_deconv}
    \widehat{\mat{A}}_{\mathrm{ilct}} = \frac{1}{\widehat{\mat{A}}_\mathrm{lct}} \frac{|\widehat{\mat{A}}_\mathrm{lct}|^2}{|\widehat{\mat{A}}_\mathrm{lct}|^2 + {1/\alpha}},
\end{align}
where $\widehat{\mat{A}}_{\mathrm{lct}} = \mat{F} \mat{A}_\mathrm{lct}$ are the Fourier coefficients of the convolution kernel, and $\alpha$ is a SNR parameter.  

\subsection{General primal-domain NLOS imaging} 
\label{sec:primal_discussion}
\input{tab_forward_inverse}
\input{tab_cont_mat}
Based on our outline of NLOS Radon transforms, in the following we show how all inverse models are special cases of a general expression for primal-domain NLOS imaging. 
The matricial form of the forward models (\cref{eq:ERT_mat,eq:PRT_mat,eq:LCT_mat}) always follows a similar expression 
\begin{equation}
    \mat{h}=\mat{A}_\square \mat{f},
\end{equation} 
where $\mat{A}\mat{}_\square$ represents a known forward operator based on the geometric constraints of the corresponding Radon transforms (\cref{eq:ERT_cont,eq:PRT_cont,eq:SRT}). To solve for $\mat{f}$, the goal is to find $\minv{\mat{A}}_\square$ so that
\begin{equation}
\label{eq:NLOS_inverse_mat}
    \mat{f}=\minv{\mat{A}}_\square \mat{h}.
\end{equation}
While this is a simpler problem than estimating $\mat{f}$ for general unconstrained transport (\cref{eq:general_formation_mat}), estimating $\minv{\mat{A}}_\square$ for constrained forward transport models (\cref{eq:ERT,eq:PRT,eq:LCT}) is still an ill-posed problem, since $\mat{A}_\square$ may not be square, it may be singular, or its inverse may be non-trivial to find. 

The imaging models that invert the ERT \cite{Velten2012nc} (\cref{eq:inv_ERT_mat}) and PRT \cite{Liu2019analysis} (\cref{eq:inv_PRT_mat}) approximate $\minv{\mat{A}}_\square$ by combining backprojection $\mat{A}_\square^\transpose$ with a filtering step $\mat{K}_\square$. 
While the approximate inversion of the SRT \cite{OToole2018confocal} (\cref{eq:inv_LCT_mat}) may seem to deviate from a traditional backprojection, note that
\begin{equation}
\label{eq:iLCT}
\frac{|\widehat{\mat{A}}_\mathrm{lct}|^2}{\widehat{\mat{A}}_\mathrm{lct}} \equiv \widehat{\mat{A}}_\mathrm{lct}^\ast,
\end{equation}
which is the Fourier-space conjugate of the forward operator $\mat{A}_\mathrm{lct}$. This corresponds to its transpose $\mat{A}_\mathrm{lct}^\transpose$ in the primal domain, namely the backprojection operator for the re-parameterized SRT (\cref{eq:LCT_mat}). \cref{eq:LCT_deconv} therefore becomes
\begin{align}
\label{eq:LCT_BP_filt}
    \widehat{\mat{A}}_{\mathrm{ilct}} = \frac{\widehat{\mat{A}}^\ast_\mathrm{lct}}{|\widehat{\mat{A}}_\mathrm{lct}|^2 + \minv{\alpha}}.
\end{align}
In the primal domain, this is again a combination of the backprojection operator $\mat{A}_\mathrm{lct}^\transpose \equiv \widehat{\mat{A}}^\ast_\mathrm{lct}$ weighted by   ${\mat{K}_\mathrm{lct} \equiv (|\widehat{\mat{A}}_\mathrm{lct}|^2+1/\alpha)^{-1}}$. We can therefore write the following equivalence for \cref{eq:inv_LCT_mat},
\begin{align}
\mat{f} & \approx \minv{\mat{R}}_z \minv{\mat{F}} \widehat{\mat{A}}_{\mathrm{ilct}} \mat{F} \mat{R}_t \mat{h},\\
\label{eq:inv_LCT_mat_BP}
  & = \minv{\mat{R}}_z \mat{K}_\mathrm{lct}  \mat{A}^\transpose_{\mathrm{lct}} \mat{R}_t \mat{h}.
\end{align}

The estimation of $\mat{f}$ for the re-parameterized SRT therefore follows the same filtered backprojection strategy as the ERT and PRT. The re-parameterized SRT, however, does provide a convenient formulation tied to confocal configurations, where the backprojection and filtering steps can be computed efficiently as a 3D Fourier-space multiplication (\cref{eq:inv_LCT}). 

This leads to the following observation: despite their differences, all the models discussed approximate the inverse of their corresponding forward operator $\mat{A}_\square$ as a combination of filtering and backprojection steps, as 

\begin{eqboxedwhite}
\begin{align}
\label{eq:inverse_as_fbp}
    \minv{\mat{A}}_\square \approx \mat{K}_\square \mat{A}_\square^\transpose.
\end{align}
\end{eqboxedwhite}
Under this homogeneous formulation, the ability to correctly estimate the inverse operator $\minv{\mat{A}}_\square$ resides exclusively on the choice of filter $\mat{K}_\square$, since $\mat{A}_\square$ and therefore $\mat{A}^\transpose_\square$ are known a priori. The choice of filter $K \equiv \mat{K}_\square$ varies across methods depending on transport assumptions or design choices. In \cref{sec:evaluation} we show how the filter shape relates to imaging constraints and reconstruction performance of the various methods.  

Based on \cref{eq:NLOS_inverse_mat,eq:inverse_as_fbp} we can write a \textbf{general expression} in continuous and matricial form for the imaging process as 
\begin{eqboxedwhite}
\begin{subequations}
\label{eq:general_FBP}
\begin{align}
\tag{\theequation\textsubscript{c}} 
\label{eq:general_FBP_cont}
\begin{split}
    \hspace{-0.5em}f(\xv)& \approx \Bigl[ K_f(\xv) \conv \iint\limits_{S\,L} \Bigl[K_H(\xl, \xs, t) \Bigr. \\
    &\Bigl. \conv H(\xs,\xl,t-(\tlv+\tsv))\Bigr] \diff \xl \diff \xs\ \Bigr]_{t=0},
\end{split}\\
\tag{\theequation\textsubscript{m}} 
\label{eq:general_FBP_mat}
    \mat{f} &\approx \mat{K}_{v,\square} \mat{A}_\square^\transpose  \mat{K}_{lst,\square} \mat{h},
\end{align}
\end{subequations}
\end{eqboxedwhite}
where $K \equiv \mat{K}$ represent filters over the domains (laser $\xl \in L$, sensor $\xs \in S$, time $t\in [0,\infty)$, scene  $\xv \in V$); and $\ast$ represents convolutions in those domains.  
\cref{tab:cont_mat_summary} summarizes the correspondence between the elements of forward and inverse models in continuous and matricial form. 

%% file: tab_forward_inverse.tex
\renewcommand{\arraystretch}{2}
\begin{table*}
\centering
\caption{Summary of forward and inverse models in NLOS imaging (\cref{sec:primal_NLOS_models}) under assumptions \ref{asm:direct} to \ref{asm:calibration}.}
\label{tab:radon_summary}
\resizebox{\linewidth}{!}{
\begin{tabular}{ c | c | c }
& \text{Forward model} & \text{Inverse model} 
\\ \hline
General 
&
&
\makecell{
    $\hspace{-0.5em}f(\xv) \approx [K_f(\xv)\conv\iint [K_H(\xl, \xs, t) $ \\ $\conv H(\xs,\xl,t-(t_{lv}+t_{sv}))] \diff \xl \diff \xs ]_{t=0}$
    } 
    (\cref{eq:general_FBP_cont})
\\ \hline
ERT 
& 
$H(\xl,\xs,t) = \bar{G} \int\delta(d_{lv}+d_{sv}-ct) f(\xv) \diff\xv$ 
(\cref{eq:ERT_cont})
&  
$f(\xv)\! \approx \nabla_{\xv}^2 \iint H(\xl,\xs,-(t_{lv}+t_{sv})) \diff \xl \diff \xs$
(\cref{eq:inv_ERT_cont})  \cite{Velten2012nc}
\\ \hline 
PRT 
& 
\makecell{$H(\xl',\xs',t-(t_{lv}+t_{sv}))\approx$ \\ $\iiint{f(\x_v')} \delta(k_x x'+ k_y y'+ k_z z') dx'dy'dz'$} 
(\cref{eq:PRT_cont})
&  
$f(\xv) \approx \iint\nabla_t^2 H(\xl,\xs,-(t_{cv}+t_{sv})) \diff \xl \diff \xs$
(\cref{eq:inv_PRT_cont})  \cite{Liu2019analysis}
\\ \hline 
SRT 
& 
$H(\xs, t)= \bar{G} \int \delta(2d_{sv}-ct) f(\xv) \diff\xv$
(\cref{eq:SRT}) 
& 
$\reparam{R}_t\{H\}(\bar\xbf_s, \bar t) = \int a(\bar\xbf_v-\bar\xbf_s) \reparam{R}_z\{f\}(\bar\xbf_v)  \diff \bar\xbf_v$ 
(\cref{eq:LCT_cont}) \cite{OToole2018confocal} 
\\  
\end{tabular}
}
\end{table*}

%% file: tab_cont_mat.tex
\renewcommand{\arraystretch}{2}
\begin{table*}
\centering
\caption{Summary of continuous and matricial form of NLOS imaging inverse models (\cref{sec:primal_NLOS_models}) under assumptions \ref{asm:direct} to \ref{asm:calibration}.}
\label{tab:cont_mat_summary}
\begin{tabular}{ c | c | c }
& \text{Continuous form} 
& \text{Matricial form} 
\\ \hline
General (\cref{eq:general_FBP})
&
\makecell{$\hspace{-0.5em}f(\xv) \approx [K_f(\xv) \conv \iint [K_H(\xl, \xs, t) \ast$ \\ $H(\xs,\xl,t-(t_{lv}+t_{sv}))] \diff \xl \diff \xs |_{t=0}$} 
&
$\mat{f} \approx \mat{K}_{\mat{f},\square} \mat{A}_\square^\transpose  \mat{K}_{\mat{h},\square} \mat{h}$
\\ \hline
ERT  \cite{Velten2012nc} (\cref{eq:inv_ERT}) 
& 
$f(\xv)\! \approx \nabla_{\xv}^2 \iint H(\xl,\xs,-(t_{lv}+t_{sv})) \diff \xl \diff \xs$
&  
$\mat{f} \approx \mat{K}_{\nabla,\xv} \mat{A}_\mathrm{ert}^\transpose \mat{h}$
\\ \hline 
PRT \cite{Liu2019analysis} (\cref{eq:inv_PRT}) 
& 
$f(\xv) \approx \iint\nabla_t^2 H(\xl,\xs,-(t_{cv}+t_{sv})) \diff \xl \diff \xs$
& 
$\mat{f} \approx \mat{A}_\mathrm{prt}^\transpose \mat{K}_{\nabla,t} \mat{h}$
\\ \hline 
SRT \cite{OToole2018confocal} (\cref{eq:inv_LCT}) 
& 
$\Fspace{f}(\freq_{\xv}) \approx \reparam{R}_{z}^{-1} \Bigl\{\Fspace{a}^{-1}(\freq_{\bar\xbf_v-\bar\xbf_s}) \reparam{R}_t\{\Fspace{H}\}(\freq_{\bar\xbf_s})\Bigr\}\left(\freq_{\xv}\right)$ 
&
$\mat{f} \approx \minv{\mat{R}}_z \minv{\mat{F}} \widehat{\mat{A}}_{\mathrm{ilct}} \mat{F} \mat{R}_t \mat{h}$
\\  
\end{tabular}
\end{table*}

%% file: src5_frequencydomain.tex
\section{Frequency-domain NLOS imaging models}
\label{sec:frequency_NLOS_models}
The phasor-field formulation \cite{Liu2019phasor} exploits wave propagation principles and existing line-of-sight optics literature to formulate existing and novel NLOS imaging models and operators \cite{Guillen2020Effect,Liu2020ICCP,Marco2021NLOSvLTM}. Moreover, since many efficient wave-based solvers for LOS imaging exist, they can be conveniently leveraged and migrated to the NLOS domain under this formulation to reduce computational costs \cite{Liu2020phasor}.
In the following we analyze the general expression of filtered backprojection methods (\cref{eq:general_FBP}) and show how such methods closely resemble specific line-of-sight propagation models under the phasor-field formulation, where the filtering functions are equivalent to virtual illumination functions, and traditional backprojection of third-bounce illumination corresponds to the virtual imaging model of a confocal camera. 

In \cref{sec:evaluation} and \ref{sec:eval_capture} we summarize the connections between performance in NLOS imaging and traditional LOS imaging principles, and experimentally demonstrate the global characteristics of the reconstructions are similar across different NLOS imaging methods in simulated and captured scenarios.

\subsection{Frequency-domain Radon Transform}
\label{sec:frequency_radon}
As shown in \cref{sec:primal_discussion}, all primal-domain solutions for {ToF} NLOS imaging involve solving a particular case of the Radon transform, and can be summarized as a filtered backprojection operation (\cref{eq:general_FBP}). 

Applying the Fourier transform on the temporal domain in \cref{eq:general_FBP} (and ignoring the evaluation at $t=0$) we obtain
\begin{equation}
\label{eq:FBP_as_RSD}
\begin{split}
    \mathcal{F}_t\left\{f(\xv, t)\right\}=\Fspace{f}(\xv, \freq) =  \\  
    \int\limits_{S} e^{i \freq t_{sv}} \int \limits_{L} e^{i \freq t_{lv}} \Fspace{K}(\xl,\xs,\freq) \Fspace{H}(\xl, \xs, \freq) \diff \xl  \diff \xs,
\end{split}
\end{equation}
where $\ \widehat{} \ $ represents the Fourier transform and $\freq$ represents frequency, and the time delays $t_{lv}$ and $t_{sv}$ depend exclusively on voxel-SPAD and voxel-laser distances.
This equation provides a general expression for time-resolved filtered backprojection in Fourier space, where $f(\xv)$ corresponds to geometry reconstructions obtained by the aforementioned inverses of different Radon transforms (\eeeref{eq:inv_ERT}{eq:inv_PRT}{eq:inv_LCT}) as
\begin{subequations}
\label{eq:inv_Fourier}
\begin{align}
\tag{\theequation\textsubscript{c}} 
\label{eq:IFFT_FBP_as_RSD_cont}
    f(\xv) & = \mathcal{F}^{-1}_t\left\{\Fspace{f}(\xv,\freq)\right\}_{t=0}, \\
\tag{\theequation\textsubscript{m}} 
\label{eq:IFFT_FBP_as_RSD_mat}
    \mat{f} &= \minv{\mat{F_t}} 
    \mat{E}_{t_{sv}}
    \mat{E}_{t_{lv}}
    \mat{F_t} \mat{K_\square} \mat{A}_\square^\transpose\mat{h},
\end{align}
\end{subequations}
with $\mat{F_t}$ the discrete Fourier transform operator in the temporal domain, and $\mat{E}_{t_{sv}}$ and $\mat{E}_{t_{lv}}$ the complex diagonal matrix scaling each voxel and frequency by the exponential terms in \cref{eq:FBP_as_RSD}. 

\citet{Liu2019phasor} highlighted the relationship between the frequency-domain version of the backprojection operator (\cref{eq:FBP_as_RSD}) and lens-based image formation models defined by Rayleigh-Sommerfeld Diffraction (RSD) propagation of complex-valued measurements (i.e. phasors) on a camera aperture. By interpreting the complex-valued $\Fspace{H}(\xl,\xs,\freq)$ as a phasor field at planar virtual apertures $L, S$ on the relay wall, a \emph{virtual} image of the hidden scene can be formed by means of RSD propagation as
\begin{align}
\label{eq:FBP_RSD_at_voxel}
    \Fspace{f}_{\textrm{cc}}(\xv, \freq) &= \int\limits_{S} \frac{e^{i \freq t_{sv}}}{d_{sv}} \Fspace{P}_s(\xs, \freq) \diff \xs, \\
\label{eq:FBP_RSD_at_SPAD}
    \Fspace{P}_s(\xs,\freq) &= \int \limits_{L} \underbrace{\frac{e^{i \freq t_{lv}}}{d_{lv}}\Fspace{K}(\xl, \xs,\freq)}_{:=\Fspace{P_l}(\xl, \xs,\freq)} \Fspace{H}(\xl,\xs,\freq) \diff \xl,
\end{align}
where $P_l$ is defined as the emission of a planar source at $L$ focused on point $\xv$, while $P_s$ is the light at $\xs$ from $P_l$ scattered from the hidden scene via the impulse response operator $H$.
Similarly, for a set of imaged locations $\xs$ on a plane $S$, $\Fspace P_s(\xs,\freq)$ (\cref{eq:FBP_RSD_at_SPAD}) represents the hidden-scene wavefront measured at $S$ under illumination $P_l$. In particular, \cref{eq:FBP_RSD_at_voxel,eq:FBP_RSD_at_SPAD} implement a \emph{virtual} confocal camera model\footnote{Note here the term \emph{confocal} refers to the co-location of virtual illumination and camera focal points, not to the topology of the captured data $H$} $\Fspace{f}_{\textrm{cc}}$ that closely resembles the implementation of a backprojection operator (\cref{eq:FBP_as_RSD}) except for the linear decay terms $1/d_{lv}, 1/d_{sv}$, which mainly affect visibility of objects depending on their distance to the relay surface, without modifying the shape.

Both $\Fspace f$ (\cref{eq:FBP_RSD_at_voxel}) and $\Fspace P_s$ (\cref{eq:FBP_RSD_at_SPAD}) are computed as superpositions of scene-dependent phase-shifted wavefronts.
These resemble ubiquitous operations in line-of-sight (LOS) imaging systems which propagate a wavefront of frequency $\freq$ from a plane $\xa \in A$ to a point $\xb$, effectively synchronizing the phase of the wavefront at $\xb$. When $\xa$ lie on an infinite plane $A$, this corresponds to to a thin-lens propagation with a planar aperture $A$. The Rayleigh-Sommerfeld Diffraction (RSD) operator performs this propagation for the particular case of propagation between two apertures $A \rightarrow B$ as
\begin{equation}
\label{eq:RSD}
\Fspace{P}_b(\xb, \freq)=\int\limits_A{\Fspace{P}_a(\xa, \freq)\frac{e^{i \freq t_{ab}}}{d_{ab}}\diff\xa}:=\mathcal {R}_{A\shortrightarrow B}(\Fspace{P}_a,\xb),
\end{equation}
where $\xa$ and $\xb$ are points in apertures $A$ and $B$, respectively, and $\mathcal {R}_{A\shortrightarrow B}$ is the RSD operator. The RSD operator has efficient closed-form solutions when solved through the frequency domains of both space $\xs \rightarrow \Omega_{\xs}$ and time $t\rightarrow \Omega_t$ \cite{Liu2019phasor,Liu2020phasor}.

\subsection{Special case: Frequency-domain SRT}
For a confocal measurement set where $\xl=\xs$, \cref{eq:FBP_as_RSD} reduces to the \emph{frequency domain SRT}, which approximates $\Fspace{f}(\xv, \freq)$ as
\begin{equation}
    \Fspace{f}(\xv, \freq) = 
    \int_{S} e^{i \freq 2 t_{sv}}  \Fspace{K}(\xs, \freq) \Fspace{H}(\xs, \freq) \diff \xs.
\end{equation}
By operating in the frequency domain of $\xs$ and $t$ in confocal setups, \citet{Lindell2019wave} efficiently solve NLOS inversion using a technique known as \emph{f-k} migration. This approach has been popular in applications where emitter and receiver are typically co-located, such as in seismology~\cite{Margrave2001,margrave_lamoureux_2019} and ultrasound~\cite{Tasinkevych2014circular} imaging. 
The \emph{f-k} migration method involves a combination of re-sampling operations and frequency-domain interpolation (known as Stolt interpolation) in a similar fashion O'Toole et al.'s efficient LCT solver  (see \cref{sec:SRT}). In contrast to the usual approximation of the inverse transport operator as a filtered backprojection operator (\cref{eq:inv_ERT}), \emph{f-k} migration implements an exact inverse operator without any filtering. However, similar to other NLOS imaging methods this inverse is based on third-bounce transport assumptions. This exact wave-based inversion is analogous to a virtual-wave forward imaging process. In \cref{sec:evaluation,sec:eval_capture} we show this method leads to similar reconstruction limitations as other NLOS imaging methods based on third-bounce assumptions. 

To support non-confocal data, Lindell et al. proposed a confocal approximation of non-confocal measurements. Note this approximation is agnostic to the confocal reconstruction method. 
In our analysis (\cref{sec:evaluation,sec:eval_capture}), we apply Lindell's approximation when dealing with non-confocal data for both LCT \cite{OToole2018confocal} and \textit{f-k} migration \cite{Lindell2019wave} so as to perform comparisons with methods that support non-confocal data \cite{Velten2012nc,Laurenzis2014feature,Liu2020phasor}. 

\subsection{General frequency-domain NLOS imaging}
Despite relying on different computational solutions, both the imaging model defined in \erref{eq:FBP_RSD_at_voxel}{eq:RSD} and \textit{f-k} migration \cite{Lindell2019wave} are based on a common strategy---posing NLOS imaging as a wave-propagation problem. Both methods image the hidden scene by syncing the phases of the Fourier-space capture function $\Fspace{H}$ at the third-bounce time-of-flight for every voxel of the hidden space. The differences between methods reside on how to solve the resulting formulae in a computationally efficient way.

The phasor field formulation imports wave propagation into NLOS imaging as combination of virtual illumination functions and RSD-like propagators. Defining NLOS imaging from a virtual-wave perspective is semantically equivalent to many LOS settings, allowing to tap into well-established optics principles and efficient solvers to perform imaging operations in NLOS spaces \cite{Liu2020phasor,nam2021low}.   
In combination with custom illumination functions, this can be used to model specific virtual imaging devices in Fourier space. For instance, under pulsed illumination functions for $\Fspace P_l$, the resulting virtual imaging model resembles a time-gated confocal camera (\cref{eq:FBP_RSD_at_voxel,eq:FBP_RSD_at_SPAD}) \footnote{Note here the term \emph{confocal} refers to the co-location of virtual illumination and camera focal points, not to the topology of the captured data $H$.}\cite{Liu2019phasor}. 
The virtual illumination function $\Fspace P_l$ is equivalent to the filter $K$ (\cref{eq:general_FBP}) used in primal-domain backprojection methods, where it has a strong influence on the imaging performance (see \cref{sec:evaluation} for further discussion). 
For instance, if the virtual illumination $\Fspace P_l \equiv \Fspace K$ represents a Laplacian filter, evaluating \cref{eq:inv_Fourier} is equivalent to the inverse of the PRT (\cref{eq:inv_PRT_cont}).
%

%% file: src6_sim_experiments.tex
\section{Evaluation of NLOS imaging performance}
\label{sec:evaluation}
In the following we perform an experimental analysis of the performance of a representative set of ToF NLOS methods on a wide range of simulated datasets. 
We generate simulated datasets using physically-based transient rendering techniques that include all illumination bounces \cite{Jarabo2014,royo2022non}.

The reconstruction quality of existing methods strongly depends on two aspects: first, the implicit or explicit filtering properties of each method; second, the capture baseline, defined by the topology and spatio-temporal resolution of the laser-SPAD targets on the relay wall, which yields a band-limited approximation of the ideal impulse response function $H$ of the hidden scene. We rely on the virtual LOS-NLOS image formation analogy (\cref{sec:frequency_NLOS_models}) to reason about the performance and limitations of existing methods based on well-known LOS imaging principles; 
in particular, we analyze sensitivity to noise, the effect of filtering, resolution accuracy, and the missing cone problem.

\subsection{Methodology of analysis}
We analyze the aforementioned problems under the following NLOS imaging methods:
\begin{itemize}
    \item {FBP-Lap}, {FBP-LoG}: Filtered backprojection \cite{Velten2012nc} using Laplacian and Laplacian-of-Gaussian spatial filtering, respectively.
    \item {LCT}: Light-cone transform \cite{OToole2018confocal}, which is constrained to confocal acquisition.   
    \item {\FK}: \emph{f-k} migration \cite{Lindell2019wave}, which is constrained to confocal acquisition.
    \item {PF-CC}: Phasor-based confocal camera model under virtual pulsed illumination functions \citet{Liu2019phasor}.
\end{itemize}

\paragraph{Implementation} 
We analyze these models in both confocal and non-confocal setups. For \FK and LCT, we apply the confocal approximation introduced by \citet{Lindell2019wave} for non-confocal data. To compute the PF-CC and FBP models under confocal setups we leverage the reparameterization step of LCT \cite{OToole2018confocal}. This yields a fast but exact frequency-domain solver for the backprojection operator. 
We compute FBP under non-confocal data with a standard iterative implementation \cite{Velten2012nc,Buttafava2015}.
We compute the PF-CC model under non-confocal data using fast diffraction-based operators introduced by \citet{Liu2020phasor}. 
To keep the analysis tractable, we fix the reconstruction parameters for all methods using the default criteria described in the respective articles and public source code. In a few representative cases, we analyze the effect of specific method parameters in the reconstruction quality.

\paragraph{Datasets and reconstruction} 
The analyzed NLOS reconstruction methods yield a scalar-valued 3D voxelized reconstruction of the hidden scene. Following standard practice, we show 2D reconstructions using maximum-intensity projections over the XYZ axis of the these voxelizations, yielding front, top, and side views of the hidden scenes. 
We set the lateral resolution of the voxelization---i.e. in the dimensions co-planar to the relay wall---to match the spatial resolution of the impulse response function. Depth resolution is method-dependent and is discussed later in this section. 
Since NLOS imaging methods differ in their output dynamic range, we normalize the values of the outputs between 0 and 1 to allow for side-by-side comparisons. 

\subsection{Quantitative evaluation}
\begin{figure*}[t!]
    \centering
    \includegraphics[width=\textwidth]{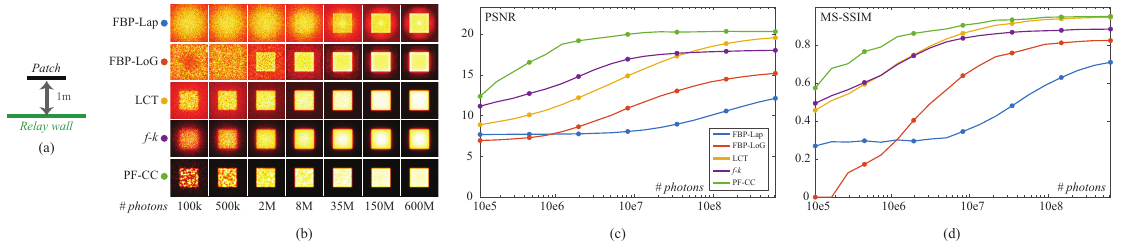}
    \caption{(a) Top view of the simulated scene made of a $\SI{1}{m}\times\SI{1}{m}$ patch co-planar and at 1 meter from the relay wall, under confocal acquisition. (b) Reconstructions obtained at increasing photon counts (columns) for every method (rows). (c,d) PSNR and MS-SSIM metrics of all the methods for the simulated photon counts ($x$ axis). Point markers correspond to the reconstructions shown in (b).}
    \label{fig:quanti_single_patch}
\end{figure*}

We first provide a quantitative evaluation of ToF NLOS imaging methods under varying noise levels. We simulate an NLOS imaging setup where the hidden scene contains a $\SI{1}{m}\times\SI{1}{m}$ patch coplanar to the relay wall, centered and at $\SI{1}{m}$ from it. We simulate a confocal capture dataset, and mimic increasing number of total captured photons by adding Poisson noise to the reference simulated dataset. \cref{fig:quanti_single_patch} shows the resulting reconstructions and error metrics for the different evaluated methods. As expected, reconstruction quality increases along with photon count, with LCT, \FK and PF-CC converging to sharp noise-free reconstructions beyond 8 million photons. LCT retrieves the patch structure at all noise levels, but suffers from background noise at lower photon counts. \FK shows sharp results beyond 2 million photons, becoming prone to high-frequency noise and loss of structure at low photon counts. PF-CC preserves the patch boundary at all photon counts, translating high-frequency photon noise into low-frequency reconstruction noise in the interior of the patch, and quickly maximizing the PSNR beyond 2M photons.

\subsection{Filtering analysis}
\label{sec:eval_noise}
Since ToF NLOS imaging systems yield data prone to noise and with limited spatio-temporal bandwidth, physically-accurate and mathematically exact inverse operators do not necessarily output optimal results. 
In turn, filtering may help to attenuate frequencies in the capture data $H$ that do not encode geometric information. 

For the original backprojection method \cite{Velten2012nc} \cref{eq:inv_ERT,eq:inv_PRT} we follow standard practice and apply Laplacian and Laplacian-of-Gaussian spatial filters over the resulting 3D reconstruction. 
LCT implements the NLOS inverse as Wiener filtering after a time-to-depth re-parameterization, providing a SNR parameter $\alpha$ (\cref{eq:LCT_BP_filt}).
\emph{\FK}-migration performs a filter-free wave-based exact inverse. 
Phasor-field virtual pulsed illumination function is defined as a time-domain Morlet wavelet---i.e. a complex exponential with carrier frequency $\lambda_c$ multiplied by a Gaussian envelope with standard deviation $\sigma$. This function acts as a filter over $H$ \emph{before} RSD propagation (\cref{eq:RSD}), preserving only a Gaussian-shaped narrow set of temporal frequencies of $\Fspace{H}$. 

To analyze the filtering properties of all methods using a common procedure, we first compute the unfiltered backprojection output $f_{\mathrm{bp}}$, and then estimate the spatial filtering properties $K_{\square}$ of each model by performing the frequency-space division between the unfiltered backprojection output $f_{\mathrm{bp}}$ and the model's output $f_\square$, 
\begin{align}
\Fspace{K}_{\square} \approx \frac{\Fspace{f}_\square}{\Fspace{f}_\mathrm{bp}}.
\label{eq:estimated_filter}
\end{align}
We highlight similarities and differences in the filtering properties of each method by visually inspecting $\Fspace{K}_{\square}$. For fair comparisons, we perform this analysis on confocal data, which is supported by all analyzed methods. 
 
\begin{figure}[t]
    \centering
    \includegraphics[width=0.95\columnwidth]{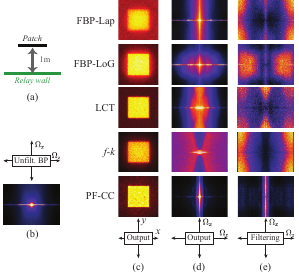}
    \caption{(a) Simulated scene with a $\SI{1}{m}\times\SI{1}{m}$ patch at 1 meter from the relay wall, confocal capture. (b) XZ frequency spectrum of the unfiltered backprojection output. (c) Front view (XY) of the reconstructions analyzed methods. (d) XZ frequency spectrum of the reconstructions yielded by each method. (e) Frequency spectrum of our estimated filtering operator for each method (\cref{eq:estimated_filter}).}
    \label{fig:filters_2D_simple}
\end{figure}
\begin{figure}[t]
    \centering
    \includegraphics[width=0.9\columnwidth]{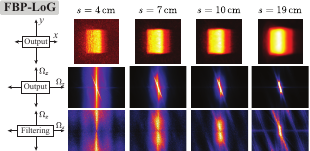}
    \caption{Variations on the width of the LoG filter, from $4$ to $\SI{19}{cm}$. Wider filters skew the reconstructions towards lower frequencies, which mitigates noise but fails to correctly reproduce surface edges.}
    \label{fig:filters_2D_LoG_param}
\end{figure}
\begin{figure}[t]
    \centering
    \includegraphics[width=0.9\columnwidth]{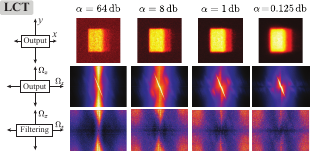}
    \caption{Variations on the LCT parameter $\alpha$ which controls response to SNR on the Wiener filtering. Lower alpha values mitigate noise at the expense of blurring the edges of the reconstructed surface.}
    \label{fig:filters_2D_LCT_param}
\end{figure}
\begin{figure}[t]
    \centering
    \includegraphics[width=0.95\columnwidth]{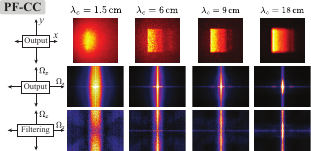}
    \caption{Variations on the central wavelength $\lambda_c = 1/\Omega_c$ for the PF-CC illumination. Increasing $\lambda_c$ mitigates noise at the expense of losing geometric detail. Wavelengths shorter than the temporal uncertainty of the system ($\SI{80}{ps} \equiv \SI{2.39}{cm}/c$) are unable to resolve the scene geometry.}
    \label{fig:filters_2D_PF_param}
\end{figure}
%
\cref{fig:filters_2D_simple}a shows a scene composed of a planar diffuse patch at one meter from the relay wall.  \cref{fig:filters_2D_simple}b shows the unfiltered backprojection output in frequency space, averaged over the $\Omega_y$ component. 
\cref{fig:filters_2D_simple}c shows a maximum-intensity projection over the $z$ dimension of the reconstruction output $\Fspace{f}_\square$ of each model. \cref{fig:filters_2D_simple}d shows the $\Omega_y$-averaged frequency spectrum of the reconstruction output $\Fspace{f}_\square$. 

\cref{fig:filters_2D_simple}e shows the estimated filtering $\Fspace{K}_\square$ (\cref{eq:estimated_filter}). 
Laplacian filtering (FBP-Lap, first row) acts as an edge detector. Due to the limitations on the hardware bandwidth, high frequencies are not captured efficiently, yielding noisy outputs. 
Laplacian-of-Gaussian (FBP-LoG, second row) provides control of the filter width, and the result is skewed towards mid-range frequencies, significantly mitigating high-frequency noise shown by the Laplacian filter. 
LCT Wiener-based filtering (third row) enhances both lower and high frequencies alike, especially in regions where there is barely any signal. 
\FK does not explicitly perform any filtering on the data---instead it implements an exact inverse, assuming all captured frequencies correspond to radiance scattered by geometry. 
Our estimation of its filtering properties using \cref{eq:estimated_filter} (fourth row) shows a spectrum skewed towards higher frequencies, akin to a Laplacian filtering (first row); this aligns with the exact PRT inverse obtained through Laplacian-filtered backprojection (\cref{sec:PRT}). 
Similarly to a Laplacian operator, \FK sensitivity to noise depends on the choice of sampling in the measurement and the reconstruction,  
which implicitly acts as a band-limiting filter. 
Last, the virtual illumination of the phasor-based confocal camera (PF-CC, last row) masks out most frequencies far from the chosen central wavelength $\lambda_c$, significantly mitigating noise. In \cref{sec:eval_resolution} we discuss the choice of this wavelength in more detail. 
Note that under total absence of noise, Laplacian-filtered backprojection and \emph{f-k} migration become optimal NLOS imaging methods. 

Several methods allow explicit parameterization of their filtering properties. 
In \cref{fig:filters_2D_LoG_param} we increase the width $s$ of the spatial LoG filtering (left to right). A wider filter skews the result towards lower frequencies (bottom row), mitigating noise while degrading geometric detail. 
\cref{fig:filters_2D_LCT_param} (left to right) decreases the $\alpha$ parameter of LCT (\cref{eq:LCT_BP_filt}), which represents the estimated SNR of the data. Low $\alpha$ values make the reconstruction more robust to noise, mitigating high lateral frequencies ($\Omega_x$) when depth frequency $\Omega_z$ is close to zero (see last row). 
\cref{fig:filters_2D_PF_param} (left to right) increases the central wavelength $\lambda_c = 1/\Omega_c$ of PF-CC, which controls the temporal frequencies that are preserved \emph{before} the backprojection step. Decreasing $\lambda_c$ provides more geometric detail, but has a lower bound where the model is unable to recover geometry. 
Laplacian filtering always computes the second derivatives at the finest data resolution available, thus has no parameters. \emph{f-k} migration is also parameter-free. In both cases, the filtering bandwidth and thereby reconstruction noise are determined by the choice of discretization in the data and reconstruction.

\subsection{Reconstruction resolution}
\label{sec:eval_resolution}
\begin{figure*}[!t]
    \centering
    \begin{subfigure}[b]{0.4859\linewidth}
        \centering
        \includegraphics[width=\linewidth]{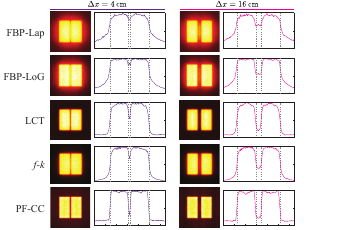}
        \caption{Fixed depth $z = \SI{1}{m}$, varying gap size $\Delta x$.}
        \label{fig:sim_gap_conf}
    \end{subfigure}
    \hspace{-2em}
    \begin{subfigure}[b]{0.514\linewidth}
        \centering
        \includegraphics[width=\linewidth]{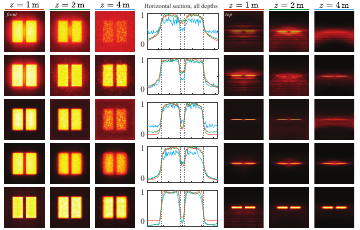}
        \caption{Fixed gap size $\Delta x = \SI{12}{cm}$, varying depth $z$.}
        \label{fig:sim_depth_conf}
    \end{subfigure}
    \vspace{-1em}
    \caption{Analysis of lateral resolution between two coplanar patches of size $\SI{0.5}{m} \times \SI{1}{m}$ under simulated confocal capture.  
    (a) Reconstructions of patches with varying gap sizes $\Delta x \in \{\SI{4}{cm}, \SI{16}{cm}\}$ at a fixed depth $z=\SI{1}{m}$. Plots show the cross sections for each method and gap size; the ground truth geometry is represented with gray dotted lines. All methods consistently fail to sharply represent small gap sizes such as $\SI{4}{cm}$ due to limitations on parameters inherent to the capture setup, such as the sampling rate of the impulse response function, aperture size, and temporal resolution. 
    (b) Reconstructions of patches at varying depths $z\in \{\SI{1}{m}, \SI{2}{m}, \SI{4}{m}\} $ with a fixed gap size of $\Delta x = \SI{12}{cm}$. Plots show the cross sections for each method, each curve represents a different depth; the ground truth geometry is represented with gray dotted lines. Consistently, all methods show increasing degradation of the lateral resolution as the distance $z$ between the relay wall and the patches increases.
    }
    \label{fig:sim_gap_depth_conf}
\end{figure*}
Next, we discuss how the parallelisms between NLOS imaging setups and LOS imaging apertures can be used to reason about the features that are reconstructed by the NLOS imaging process, and how to choose filtering properties given the limitations of the imaging systems.

In typical laboratory NLOS setups the imaging hardware is set up close to the relay wall, which allows to capture points on such relay wall with very high spatial resolution. NLOS imaging resolution is then mainly governed by the temporal jittering of the detector and the acquisition topology, which are independent of the reconstruction model. The spatial or temporal filtering properties specific to each reconstruction model may also influence the minimum reconstruction resolution of that model.

The {depth} resolution $\Delta z$ of an NLOS imaging system refers to the accuracy to measure the position of a single surface along the depth axis $z$ of the reconstruction. Larger $\Delta z$ results in less depth accuracy.
$\Delta z$ is linearly related to the temporal resolution of the detection system and can be estimated as \cite{OToole2018confocal}
\begin{equation}
\Delta z \geq \frac{c \gamma}{2},
    \label{eq:fwhm_depth}
\end{equation}
where $\gamma$ is the full width at half maximum (FWHM) of the detection system. For Gaussian-like jittering with standard deviation $\sigma$, $\gamma = 2 \sqrt{2 \ln{2}} \sigma$ and does not depend on depth.

The lateral resolution $\Delta x$ refers to the minimum resolvable distance between two scatterers that lie on a plane co-planar to the relay wall. For NLOS imaging, \citet{OToole2018confocal} estimate this lateral resolution as 
\begin{equation}
    \Delta x \geq \frac{\sqrt{(D_\textrm{max}/2)^2 + z^2}}{D_\textrm{max}} c \gamma,
    \label{eq:fwhm_lateral_full}
\end{equation}
where $D_\textrm{max}$ is the maximum distance between illuminated and sensed points on the relay wall. For typical NLOS imaging systems where $z\gg D_\textrm{max}$, $\Delta x$ increases linearly with respect to depth as
\begin{equation}
   \Delta x \geq c \gamma  \frac{z}{D_\textrm{max}} \given[\Big] z\gg D_\textrm{max}.
   \label{eq:fwhm_lateral}
\end{equation}
Note how the ability to resolve lateral details degrades as hidden objects move away from the relay wall. 

Alternatively, Buttafava et al. \cite{Buttafava2015} compared the NLOS imaging system to a camera aperture to estimate spatial resolution. The spatial resolution $\Delta x$ of a regular imaging system with imaging wavelength $\lambda$, focal length $f$, and aperture $A$ is defined by the Rayleigh criterion \cite{Rayleigh1879} as $\Delta x = 1.22 \lambda f/A$. Relating focal length to depth ${f \equiv z}$, and the aperture to the maximum light-detector distance ${A \equiv D_\textrm{max}}$, the minimum lateral resolution $\Delta x$ of an NLOS imaging system can be estimated as 
\begin{equation} 
    \Delta x \geq 1.22 c \tau \frac{z}{D_\textrm{max}},
    \label{eq:rayleigh_lateral}
\end{equation}
where the temporal uncertainty of the detector system $\tau$ establishes a lower bound of the minimum non-ambiguous temporal wavelength $\lambda \equiv \tau$ at the detector.

Note that \cref{eq:rayleigh_lateral} and \cref{eq:fwhm_lateral} both define the lower bound for lateral resolution as increasing linearly with temporal resolution $\gamma \approx 1.22 \tau$ and depth $z$, and decreasing linearly with aperture size $D_\textrm{max}$. 
As per \cref{eq:fwhm_lateral_full}, lateral resolution converges to the depth resolution of the system $\Delta x \equiv \Delta z =  c \gamma / 2$ near the wall $z\to 0$. 
Therefore, any temporal wavelength below this threshold will not provide reliable geometric information. 

While discussed in works prior to wave-based approaches, these resolution principles agree with recent methods that address NLOS imaging as a wave-based problem. 
The pulsed virtual illumination used in phasor field methods \cite{Liu2019phasor,Liu2020phasor,Marco2021NLOSvLTM} is typically defined based on this resolution limit. In practice, filtering out wavelengths shorter than four times the system jitter $\lambda_\textrm{min} = 4 \gamma$ provides a good trade-off between definition and sensitivity to noise. Wavelengths shorter than this cutoff are captured very inefficiently by the hardware, and including them usually contributes to noise in the reconstruction. \cref{fig:filters_2D_PF_param}, top row, shows this limit with $\lambda_\textrm{min} = \SI{6}{cm}$. Relying on wavelengths much larger than $\lambda_\textrm{min}$ provides spatial averaging or smoothing of the reconstruction. This is a very general way to trade resolution for noise in any data. 

The sampling spacing of the points on the relay wall and the size of the sampled spots may also constrain the imaging resolution. For wavefront shaping, it is typically assumed that a wavefront can be generated or measured when detector points are a quarter wavelength apart \cite{wiki:Phased_array}. 
A different constraint can be derived from the Nyquist limit stating that the sampling should be at least twice the highest frequency or half the wavelength.
With time resolutions of 50 to 80 picoseconds and 1-2 meter large relay walls, the imaging systems evaluated here achieve resolutions of a few centimeters at two meters from the relay wall. 

We study spatial resolution in NLOS scenes by analyzing the ability of each reconstruction model to reproduce a lateral gap between two patches. 
In \cref{fig:sim_gap_conf}, we show front views of the two patches at one meter depth with two gap sizes $\Delta x = \SI{4}{cm}$ and $\Delta x = \SI{16}{cm}$. Plots show the corresponding cross sections and the ground truth (gray lines). Small lateral gaps are challenging to reproduce for all methods due to limitations of the capture configuration, such as the sampling rate at the relay wall or the aperture size. 

In \cref{fig:sim_depth_conf} we show reconstructions of a simulated confocal scene with two $\SI{0.5}{m}\times\SI{1}{m}$ co-planar patches at varying depths of 1, 2 and 4 meters from the relay wall, with gap size fixed at \SI{12}{cm}. Reconstructions show how lateral resolution degrades as depth increases, regardless of the reconstruction model. This effect is also observed in the cross section plots (fourth column) compared to the ground truth (dotted gray lines). Some models such as FBP-Lap, FBP-LoG, or LCT enhance higher frequencies, but degrade the reconstructions at larger depths with low SNR. \FK is sensitive to noise and blurs the gap between patches, specially at larger depths. Overall, increasing depth introduces a trade-off between reconstructing blurry edges and noise level. PF-CC mitigates noise at larger depths using a narrow-band virtual illumination function that masks out higher frequencies. 

Theoretical depth resolution $\Delta z$ of the imaging setup is constant along depth (\cref{eq:fwhm_depth}), but the filtering properties of each model may determine a higher minimum resolution above that value.
We illustrate this by computing top views of our two-patch scene in \cref{fig:sim_depth_conf} (last three columns). PF-CC blurs out depth details due to the width of the virtual illumination pulse, but provides consistent low-noise results regardless of the distnace of the patches from the relay wall. Other methods (e.g. LCT, \FK) may provide sharper depth details at the expense of amplifying noise when the geometry is further from the relay wall. 
\subsection{Visibility of surfaces}
A pathological issue in NLOS imaging reconstructions is the difficulty to reconstruct planar surfaces at particular positions and angles with respect to the relay wall, even when they scatter third-bounce indirect illumination towards the relay wall. 
This effect is sometimes attributed to Lambertian shading, but results in a stronger falloff than the gradual cosine-weighted falloff expected from Lambertian shading. This is known as the missing cone problem, and it is not exclusive to NLOS imaging, but also affects other imaging applications \cite{Benning2015,Delaney1998,Mertz2019,Lim2015}.
Multiple works provided insights on the missing cone problem in NLOS imaging. 
Liu et al.~\cite{Liu2019analysis} performed a frequency analysis of the problem based on the Fourier slice theorem, showing the NLOS measurement space does not capture information of surfaces at certain orientations, showing a so-called \emph{missing cone} of frequencies in the frequency spectrum of $H$. Recent work by \citet{pueyociutad2024polNLOS} incorporated polarization into third-bounce ToF NLOS imaging, dramatically mitigating the missing cone problem thanks to directional information encoded in the polarization state of light.
\citet{royo2023virtual} relied on the virtual-wave analogy of NLOS imaging to reason about visibility in NLOS imaging. They showed planar diffuse surfaces exhibit mirror-like reflectance when imaged using third-bounce NLOS imaging assumptions, demonstrating any point in a planar surface will only be visible if it forms a specular path between the laser position and the sensor domain. 

\begin{figure}[t!]
    \centering
    \includegraphics[width=0.95\columnwidth]{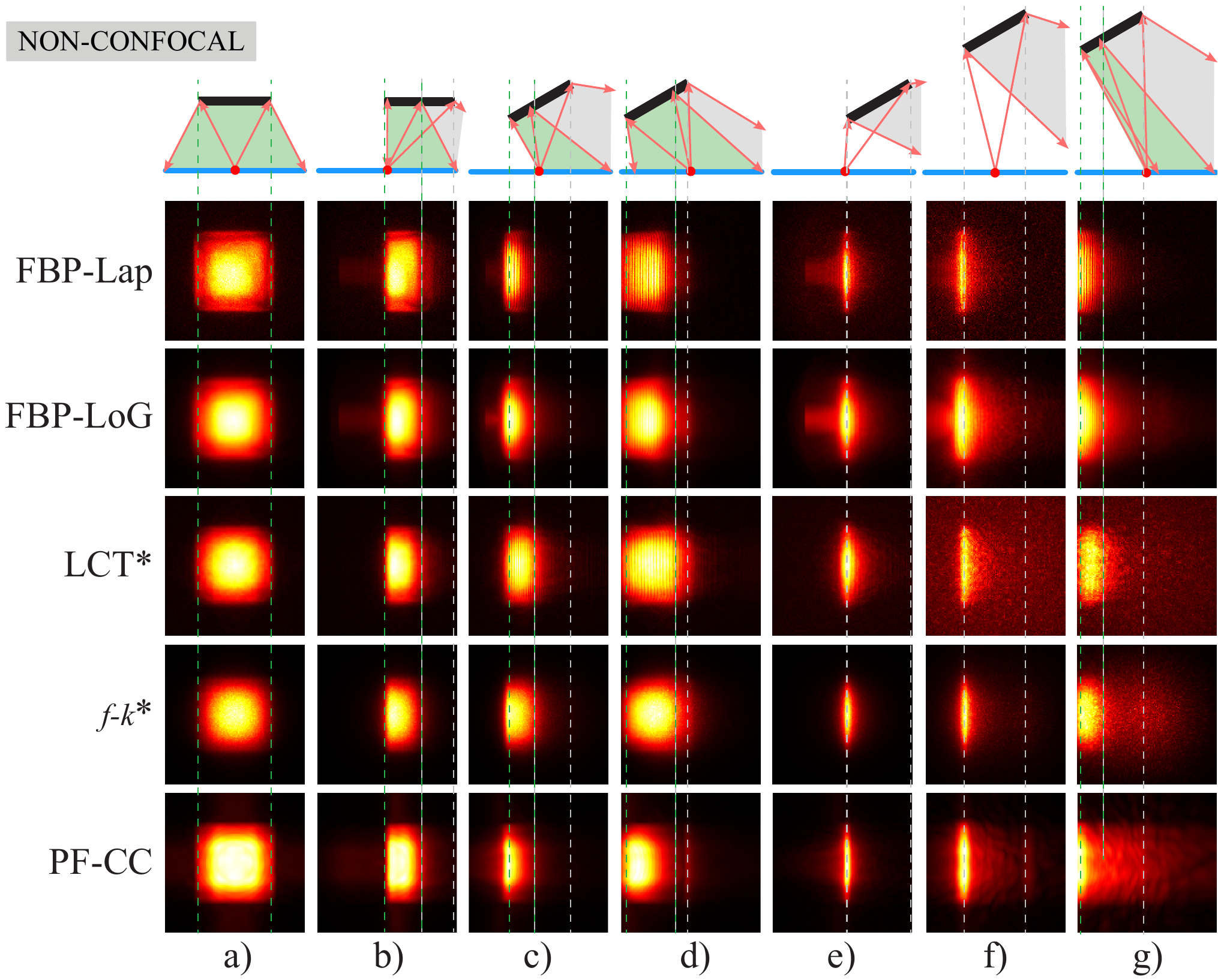}
    \caption{Analysis of surface visibility under non-confocal acquisition. First row shows a top-view schematic of scenes with a single patch under position and rotation changes. Red arrows represent light paths starting at the laser point and following a virtual specular reflection. Green and gray regions denote the area of the patch (denoted with dotted lines) that hits or misses, respectively, the SPAD domain (blue) after virtual specular reflection. a, b) Visibility of co-planar patches depends on the patch lateral position. c-f) Lateral offset, rotation, and depth jointly decrease visibility. }
    \label{fig:sim_patch_visibility_nconf}
\end{figure}
\begin{figure}[t!]
    \centering
    \includegraphics[width=0.95\columnwidth]{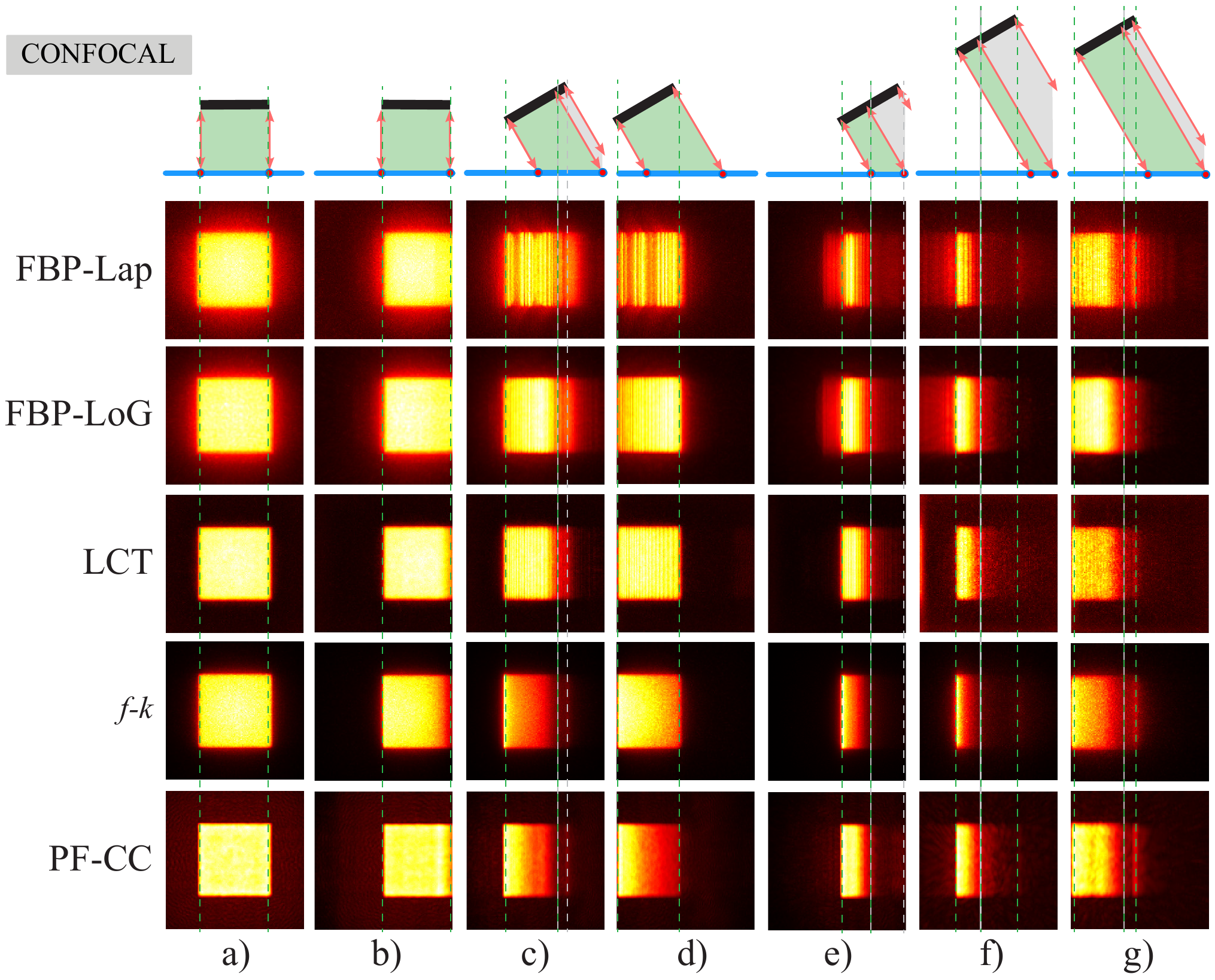}
    \caption{Analysis of surface visibility under confocal acquisition in the same scene setups as \cref{fig:sim_patch_visibility_nconf}. A surface point will be visible when its normal intersects with the confocal SPAD-laser domain. This yields higher visibility of non-coplanar and surfaces in peripheral regions.}
    \label{fig:sim_patch_visibility_conf}
\end{figure}
Under the virtual-wave analogy, it is important to account for the scale of the imaging wavelengths with respect to the scale of the geometric features that will be imaged. 
The temporal uncertainty of an NLOS imaging system (\cref{sec:eval_resolution}) usually determines the smallest theoretical wavelength which can be used to image geometry non-ambiguously. 
Under the usual temporal uncertainty of state-of-the-art ToF NLOS imaging capture setups, this wavelength is usually between $4$ to $\SI{10}{cm}$, which can be much larger or smaller than some geometric features of the hidden scene. 
By optics principles, surfaces with geometric features regarded as planar compared to the wavelength of an imaging system---e.g. a $\SI{20}{\centi\metre}\times\SI{20}{\centi\metre}$ patch under a $\SI{5}{cm}$ wavelength---will reflect light specularly and will be perceived as a mirror by the imaging system. 
NLOS imaging follows this principle \cite{royo2023virtual}, and planar hidden surfaces may become partially or completely invisible to the NLOS imaging output depending on their orientation and position with respect to the emitter and sensor baselines at the relay wall. 
While we could perform NLOS imaging at larger wavelengths of $H$ that do not interact specularly with such surfaces, lateral resolution of our imaging system is proportional to the imaging wavelength (\cref{eq:rayleigh_lateral}). 
Larger wavelengths may allow us to image surfaces that are invisible to shorter wavelengths, but at the expense of losing lateral resolution. 

We experimentally show NLOS imaging methods based on third-bounce assumptions share the same visibility issues, with minor differences that stem from their specific filtering properties.
\cref{fig:sim_patch_visibility_nconf,fig:sim_patch_visibility_conf} illustrate several cases of visibility by placing a planar patch in front of the relay wall under changes on depth, lateral position, and vertical rotations. 
In both figures, the first row shows top-view schematics of the setup, and the corresponding specular paths between the laser (red) the patch (black) and the sensor domain (blue)---green regions indicate the domain of the patch which generates specular paths reaching the sensor domain; gray regions indicate reflections missing the sensor domain.
The remaining rows show the outputs of each evaluated method. 
Dotted lines highlight the boundaries of the visible and invisible regions of the patch on the outputs. 
\cref{fig:sim_patch_visibility_nconf} shows a non-confocal setup, with the laser position (red dot) at the center of the sensor domain (blue).
\cref{fig:sim_patch_visibility_nconf}a and b show a coplanar patch whose visibility depends on the lateral position---regions to the right of the aperture center cannot be reconstructed (\cref{fig:sim_patch_visibility_nconf}b). 
\cref{fig:sim_patch_visibility_nconf}c, d and e show a patch rotated 30 degrees to the right, at different lateral offsets---the patch is more visible when moved to the left, away from the rotation direction. 
\cref{fig:sim_patch_visibility_nconf}c and f compare the same rotation and lateral offset, but different depth---increasing depth on a rotated patch decreases visibility. 

In the confocal setup (\cref{fig:sim_patch_visibility_conf}) each sensor position measures indirect light from a laser pointed at the same location as the sensor. 
This introduces a special case of surface visibility---a point on a planar surface will be visible to the imaging system only when the surface normal at that point points at the measurement domain. 
Differently from non-confocal, a surface coplanar to the relay wall will be visible at any lateral position in front of it (\cref{fig:sim_patch_visibility_conf}a, b). For non-coplanar surfaces (\cref{fig:sim_patch_visibility_conf}b-g) visibility decreases when moving in the direction of the rotation (e) or far from the wall (f, g). 
This is similar to non-confocal setups, while confocal setups have better visibility performance.
However, as we show in our real experiments, confocal setups they require longer captures gather data with the same SNR as non-confocal captures. 

%% file: src7_real_experiments.tex
\section{Evaluation in real scenes}
\label{sec:eval_capture}
In the following we illustrate our analysis under real capture setups with confocal and non-confocal acquisition topologies. 
For non-confocal setups we follow the acquisition procedure of \citet{nam2021low} by focusing two 1D SPAD arrays of $1\times14$ pixels \cite{Renna2020fast} at a fixed small area on the relay wall, which allows us to gather a higher number of photons than using single-pixel sensors under the same capture time. For confocal captures we use a single-pixel SPAD sensor, and scan a set of co-located laser and sensor points on the relay wall. 
Next, we analyze the null reconstruction space across methods and acquisition topologies, on scenes with patch distances, lateral positions, and orientations. 

\paragraph{Capture setup}
Our capture system is equipped with an ultra-fast pulsed laser (OneFive Katana HP) that operates at the wavelength of 532 nm at 0.7 W. The galvanometer (Thorlabs GVS012) drives the laser illumination and scans the relay wall. We capture confocal data using a single-pixel SPAD co-located with the laser positions. We capture non-confocal data using two 1D SPAD arrays of 14 pixels each concatenated horizontally. We implement a sparse capture procedure \cite{nam2021low} by illuminating only 1 every 28 laser positions horizontally and measuring the response with the 1x28 SPAD array, which provides a good approximation of a dense array of lasers measured by a single SPAD pixel targeted at the center of the SPAD arrays. Our SPAD arrays and single-pixel SPAD are connected to 8-channel PicoQuant HydraHarp 400 Time-Correlated Single Photon Counting (TCSPC) device. The temporal uncertainty of the system is approximately 70-90 ps. 

\subsection{Photon count}
First, we analyze performance under variations of total photon count in both confocal and non-confocal setups. 
To provide equal-photon comparisons, we manually set to zero the initial temporal bins of the captured impulse response function for both modalities to ensure the total photon count corresponds to indirect illumination in both modalities. 
Nevertheless, note that matching the total photon count between the two capture topologies is challenging due to direct photons contributing to most of the signal in confocal setups.
For confocal data we lack datasets beyond 5 million photons (two last rows), as confocal setups require much higher acquisition times than non-confocal setups to gather a similar number of photons due to the use of single-pixel SPAD sensors. 
For reference, non-confocal acquisition takes only 0.2 seconds to gather 0.5M photons in scenes with a $\SI{17}{cm} \times \SI{8}{cm}$ co-planar and centered diffuse white patch at one meter from the relay wall, while confocal acquisition takes 10 seconds to gather a similar photon count. This is consistent with our use of $1\times 28$ SPAD arrays in non-confocal setups in contrast to single-pixel confocal acquisition, which speeds up non-confocal photon gathering by almost an order of magnitude.

\begin{figure*}[t!]
    \centering
        \includegraphics[width=0.9\linewidth]{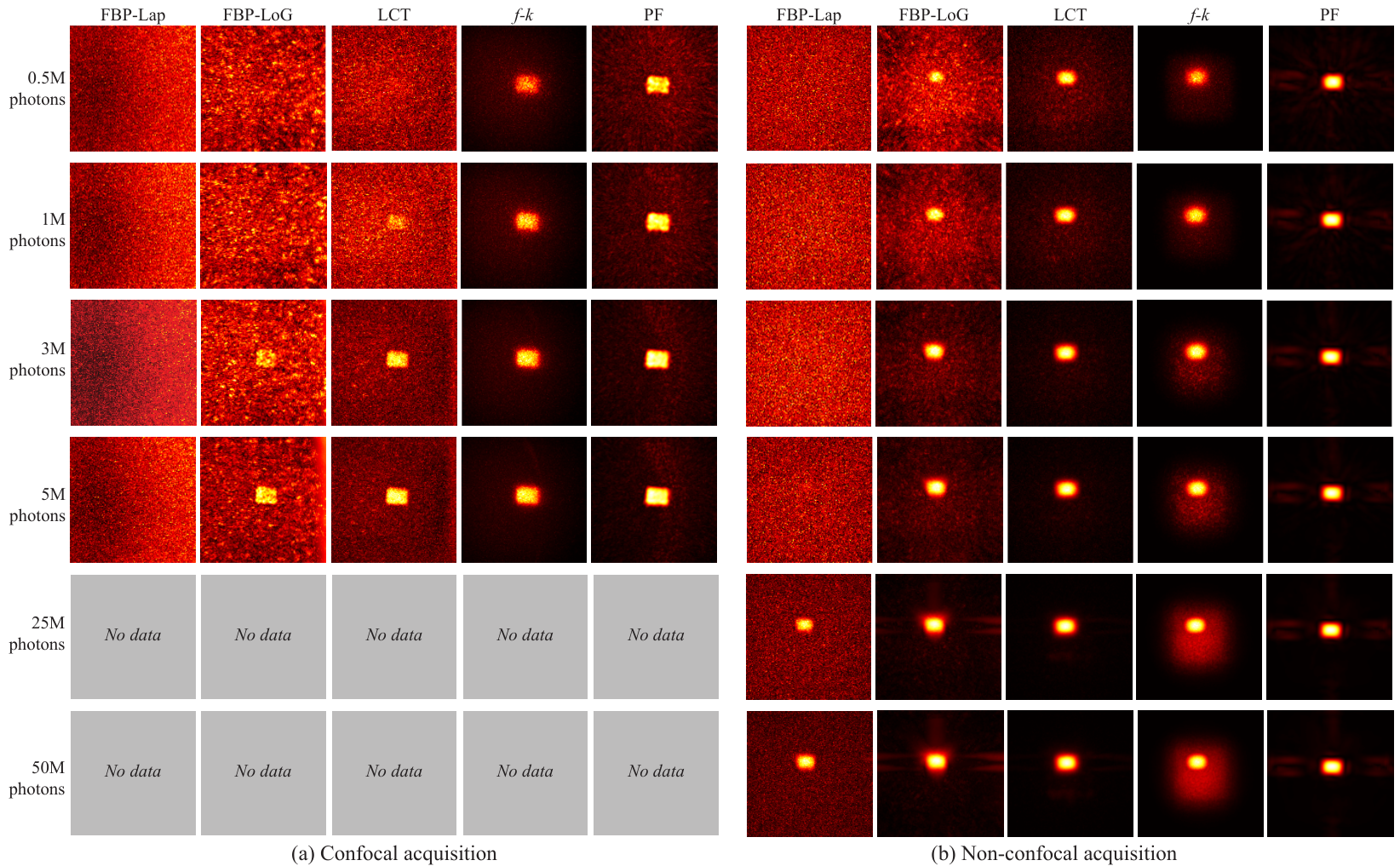}
     \vspace{-0.5em}
    \caption{Comparison of reconstructions of a single centered patch at 1 meter captured under equal-photon counts (rows) in confocal (left table) and non-confocal setups (right table). Note no data is available for 25M and 50M photon counts for confocal data (left column, two last rows). }
    \label{fig:cap_exposure}
\end{figure*}

In \cref{fig:cap_exposure} we show reconstruction results of a single $\SI{17}{cm} \times \SI{8}{cm}$ patch coplanar and centered w.r.t. to the relay wall, placed at one meter depth, captured with confocal (a) and non-confocal topologies (b). Each column shows the front view of the reconstructions obtained for each tested method. Rows show variations on the total number of photons gathered by the capture device (from 0.5M  to 50M photons). 
Under confocal data (\cref{fig:cap_exposure}a), LoG and Laplacian filtering are sensitive to noise, even at high photon counts. LCT performs better but still showing noise. 
Conversely, non-confocal data (\cref{fig:cap_exposure}b shows better performance than confocal data for most of the methods as the total photon count increases. 
Only Laplacian filter shows significant noise even at high photon counts (50M, last row).  
The confocal approximation for non-confocal data shows good results for LCT, which only shows noticeable noise under 3M photons. 
\fkmig shows a uniform component at the central part of the reconstruction which becomes more noticeable as the photon count increases. 
PF-CC (last column) shows consistent clean results regardless of the number of photons, becoming robust to noise with as few as 0.5M photons (first row). 

\subsection{Surface visibility}
\begin{figure}[t!]
    \centering
        \includegraphics[width=\columnwidth]{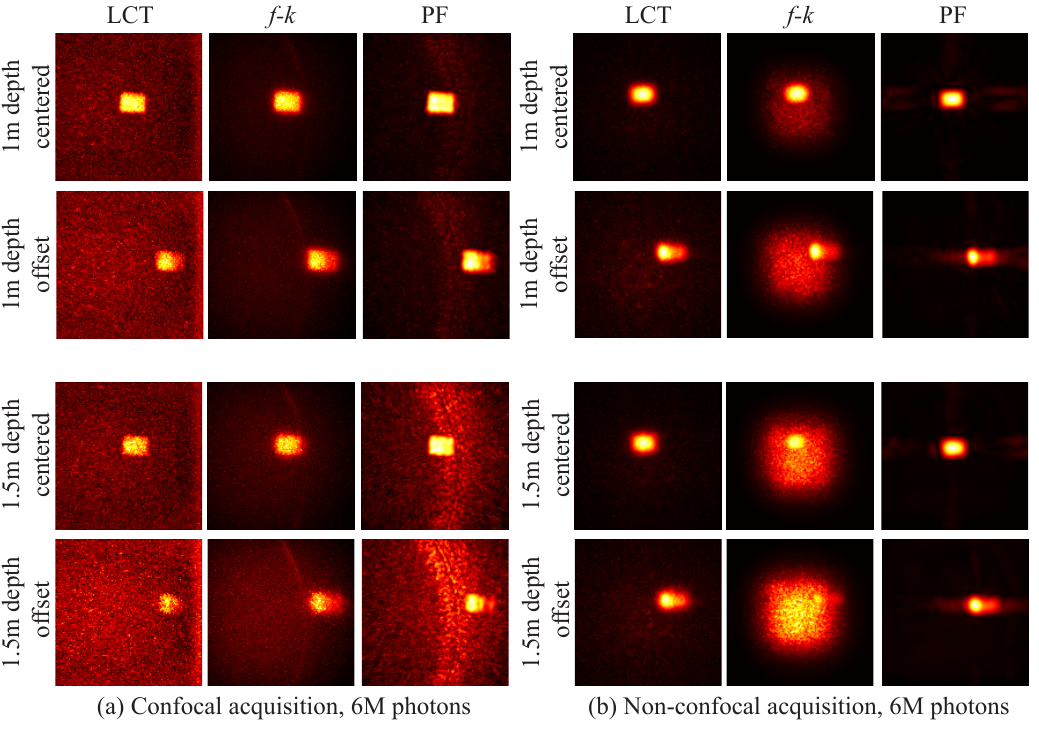}
    \caption{Comparison of reconstructions of a single patch captured under variations on depth and horizontal offset, under confocal (a) and non-confocal setups (b) with approximately 6M gathered photons.}
    \label{fig:cap_depth_lat}
\end{figure}
In \cref{fig:cap_depth_lat} we replicate our depth and lateral offset experiments in real setups using a single patch in confocal (a) and non-confocal setups (b) gathering approximately 6M photons. Consistent with our simulation experiments, confocal reconstructions with real data show better visibility properties than non-confocal reconstructions across all methods when the patch moves away from the center of the relay wall. However, LCT and PF-CC become much more prone to noise under confocal data, despite gathering a similar photon count than non-confocal data capture noise in the reconstructions. \fkmig performs better under confocal data, while the aforementioned constant component dominates the center of the reconstruction as the patch moves away laterally and in depth from the relay wall. Note reconstructing a small patch is less prone to show the missing cone problem, which is more noticeable when imaging large planar surfaces with dimensions closer to meter scale. 

\begin{figure*}[t!]
    \centering
    \includegraphics[width=0.9\linewidth]{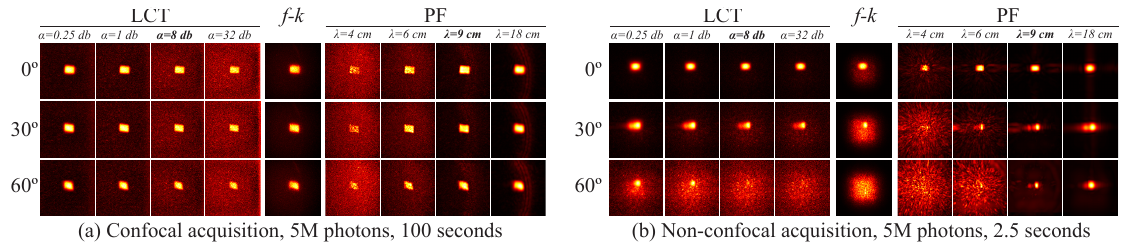}
    \vspace{-1em}
    \caption{Comparison of a single centered patch at 1m depth under variations in vertical rotation of 0, 30, and 60 degrees (rows) in confocal (a) and non-confocal setups (b) gathering approximately 5 million number of photons in total. LCT and PF tables show reconstructions for increasing parameter values: LCT varies the SNR parameter $\alpha$; PF varies the carrier wavelength $\lambda_c$ of the  pulsed virtual illumination function.  Default method parameter values are highlighted in bold.}
    \label{fig:cap_rotation}
\end{figure*}
In \cref{fig:cap_rotation} we show performance of LCT, \FK, and PF-CC under changes in vertical rotation of a planar patch at 1 meter from the relay wall, under confocal and non-confocal data. We test different values for the $\alpha$ parameter in LCT and $\lambda_c$ in PF-CC. As the patch rotates away from the relay wall (second and third rows), the sensor gathers less indirect photons from the patch. Both the $\alpha$ parameter in LCT and the central wavelength $\lambda_c$ in PF-CC provide a tradeoff between noise and reconstruction definition, requiring low $\alpha$ values and high $\lambda_c$ values, respectively, to mitigate noise and reproduce the patch as it faces away from the relay wall.

\subsection{Lateral resolution}
\begin{figure*}[t!]
    \centering
    \includegraphics[width=0.9\linewidth]{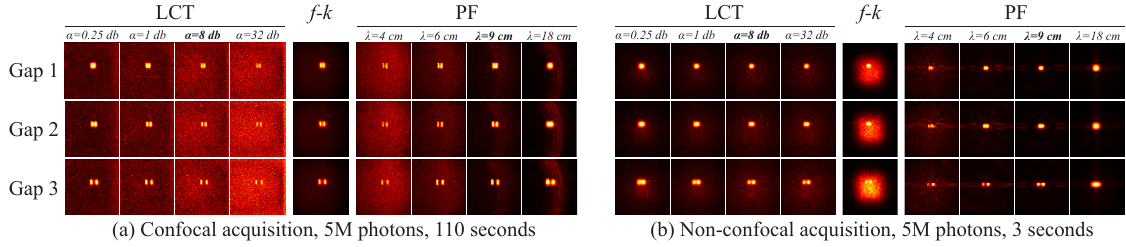}
    \vspace{-1em}
    \caption{Comparison of two patches at 1m depth with varying gap sizes between them (rows) in confocal (a) and non-confocal setups (b) gathering approximately 5 million number of photons in total. LCT and PF tables show reconstructions for increasing parameter values: LCT varies the SNR parameter $\alpha$; PF varies the carrier wavelength $\lambda_c$ of the  pulsed virtual illumination function.  Default method parameter values are highlighted in bold.}
    \label{fig:cap_lateral_gap}
\end{figure*}
In \cref{fig:cap_lateral_gap} we perform an analogous parameter exploration as in \cref{fig:cap_rotation}, but placing two side-by-side patches in front of the relay wall, and analyze performance under varying horizontal gap sizes. The parameters $\alpha$ (LCT) and $\lambda_c$ (PF-CC) determine the ability to reproduce the gap between the patches and the sensitivity to high-frequency noise, with the smallest gap size (first row) only visible in confocal data at the highest $\alpha$ (LCT) and lowest $\lambda_c$ (PF-CC) values, at the expense of introducing noise. While \fkmig performs well under confocal data, the gap is visibly less defined than in LCT or PF-CC for all gap sizes. 

%% file: src8_conclusions.tex
\section{Conclusions}
\label{sec:conclusions}
Research on ToF NLOS imaging has fostered several methods with heterogeneous formulae, implemented on multiple capture setups that yield input data of different characteristics, and showcased under scene, capture, and imaging parameters that are often carefully selected to illustrate method-specific benefits. 
As a consequence, it becomes difficult for researchers to acquire comprehensive understanding across methods.

We have carried out a methodological study of ToF NLOS imaging on a representative subset of confocal and non-confocal approaches, of which we provide a comprehensive analysis of both theoretical and experimental aspects.
We have formulated a common problem statement for ToF NLOS imaging based on the transient path integral formulation, deriving a general forward model that describes light transport in hidden scenes in continuous and matricial form, and summarized the common assumptions made by existing methods over this model (\cref{sec:problem_statement}). 

We have discussed simplified forward models proposed in the literature as specific instances of our general model, along with their corresponding inverse models (\cref{sec:primal_NLOS_models}). 
We have shown how these specific models are inherently different kinds of Radon transforms, which motivate the use of efficient solvers or specific filtering strategies. From these, we have formulated a primal-domain generic expression that encompasses NLOS inverse models proposed by several methods under typical light transport assumptions.
We have outlined the different models in both continuous and matricial forms, as they are interchangeably used in the NLOS imaging literature.
We have discussed how the frequency-domain counterpart of the general expression for NLOS imaging relates to the phasor-based formulation \cite{Liu2019analysis}, which introduced analogies between wave-based LOS image formation using RSD propagation and the reconstruction principles of ToF NLOS imaging (\cref{sec:frequency_NLOS_models}).

We have performed an experimental evaluation of different methods with the same datasets, both simulated and captured in calibrated scenarios, and exploit the analogy between wave-based LOS image formation and NLOS imaging to perform an analysis of their frequency spectrum, reconstruction noise, spatial resolution, and surface visibility in hidden scenes (\cref{sec:evaluation,sec:eval_capture}). We have shown that, when operating under analogous capture conditions, existing NLOS imaging methods share similar performance limitations, since they share common theoretical principles. Particular differences on performance between methods stem from method-specific parameters, mitigating negative features such as noise at the expense of degrading desired features such as resolution. Our analysis using captured data shows analogous deterioration of performance across all methods due to sparse and imperfect sampling of the imaged domain.
We expect that our analysis and data will become useful for future research on ToF NLOS imaging, setting theoretical and experimental grounds for novel methods on this field.

\ifDataAvailability
\section*{Data availability}
No publicly-available datasets were used in the development of this manuscript.
\fi

%% file: main.bbl

\begin{thebibliography}{50}


\ifx \showCODEN    \undefined \def \showCODEN     #1{\unskip}     \fi
\ifx \showISBNx    \undefined \def \showISBNx     #1{\unskip}     \fi
\ifx \showISBNxiii \undefined \def \showISBNxiii  #1{\unskip}     \fi
\ifx \showISSN     \undefined \def \showISSN      #1{\unskip}     \fi
\ifx \showLCCN     \undefined \def \showLCCN      #1{\unskip}     \fi
\ifx \shownote     \undefined \def \shownote      #1{#1}          \fi
\ifx \showarticletitle \undefined \def \showarticletitle #1{#1}   \fi
\ifx \showURL      \undefined \def \showURL       {\relax}        \fi
\providecommand\bibfield[2]{#2}
\providecommand\bibinfo[2]{#2}
\providecommand\natexlab[1]{#1}
\providecommand\showeprint[2][]{arXiv:#2}

\bibitem[Ahn et~al\mbox{.}(2019)]%
        {ahn2019convolutional}
\bibfield{author}{\bibinfo{person}{Byeongjoo Ahn}, \bibinfo{person}{Akshat
  Dave}, \bibinfo{person}{Ashok Veeraraghavan}, \bibinfo{person}{Ioannis
  Gkioulekas}, {and} \bibinfo{person}{Aswin~C Sankaranarayanan}.}
  \bibinfo{year}{2019}\natexlab{}.
\newblock \showarticletitle{Convolutional approximations to the general
  non-line-of-sight imaging operator}. In \bibinfo{booktitle}{\emph{Proc. of
  the IEEE/CVF International Conference on Computer Vision}}.
  \bibinfo{pages}{7889--7899}.
\newblock


\bibitem[Benning et~al\mbox{.}(2015)]%
        {Benning2015}
\bibfield{author}{\bibinfo{person}{Martin Benning}, \bibinfo{person}{Christoph
  Brune}, \bibinfo{person}{Marinus~Jan Lagerwerf}, {and}
  \bibinfo{person}{Carola-Bibliane Sch{\"o}nlieb}.}
  \bibinfo{year}{2015}\natexlab{}.
\newblock \showarticletitle{TGV sinogram inpainting for limited angle
  tomography}.
\newblock \bibinfo{journal}{\emph{Proc. of the Royal Society A}}
  (\bibinfo{year}{2015}).
\newblock


\bibitem[Bouman et~al\mbox{.}(2017)]%
        {bouman2017turning}
\bibfield{author}{\bibinfo{person}{Katherine~L Bouman}, \bibinfo{person}{Vickie
  Ye}, \bibinfo{person}{Adam~B Yedidia}, \bibinfo{person}{Fr{\'e}do Durand},
  \bibinfo{person}{Gregory~W Wornell}, \bibinfo{person}{Antonio Torralba},
  {and} \bibinfo{person}{William~T Freeman}.} \bibinfo{year}{2017}\natexlab{}.
\newblock \showarticletitle{Turning corners into cameras: Principles and
  methods}. In \bibinfo{booktitle}{\emph{Proc. of the IEEE International
  Conference on Computer Vision}}. \bibinfo{pages}{2270--2278}.
\newblock


\bibitem[Buttafava et~al\mbox{.}(2015)]%
        {Buttafava2015}
\bibfield{author}{\bibinfo{person}{Mauro Buttafava}, \bibinfo{person}{Jessica
  Zeman}, \bibinfo{person}{Alberto Tosi}, \bibinfo{person}{Kevin Eliceiri},
  {and} \bibinfo{person}{Andreas Velten}.} \bibinfo{year}{2015}\natexlab{}.
\newblock \showarticletitle{Non-line-of-sight imaging using a time-gated single
  photon avalanche diode}.
\newblock \bibinfo{journal}{\emph{Opt. Express}} \bibinfo{volume}{23},
  \bibinfo{number}{16} (\bibinfo{year}{2015}).
\newblock


\bibitem[Chen et~al\mbox{.}(2019)]%
        {Chen2019steady}
\bibfield{author}{\bibinfo{person}{Wenzheng Chen}, \bibinfo{person}{Simon
  Daneau}, \bibinfo{person}{Fahim Mannan}, {and} \bibinfo{person}{Felix
  Heide}.} \bibinfo{year}{2019}\natexlab{}.
\newblock \showarticletitle{Steady-state non-line-of-sight imaging}. In
  \bibinfo{booktitle}{\emph{Proc. of the IEEE/CVF Conference on Computer Vision
  and Pattern Recognition}}. \bibinfo{pages}{6790--6799}.
\newblock


\bibitem[Cohen(1982)]%
        {cohen1982anti}
\bibfield{author}{\bibinfo{person}{Adam~Lloyd Cohen}.}
  \bibinfo{year}{1982}\natexlab{}.
\newblock \showarticletitle{Anti-pinhole imaging}.
\newblock \bibinfo{journal}{\emph{Optica Acta: International Journal of
  Optics}} \bibinfo{volume}{29}, \bibinfo{number}{1} (\bibinfo{year}{1982}),
  \bibinfo{pages}{63--67}.
\newblock


\bibitem[Deans(2007)]%
        {deans2007radon}
\bibfield{author}{\bibinfo{person}{Stanley~R Deans}.}
  \bibinfo{year}{2007}\natexlab{}.
\newblock \bibinfo{booktitle}{\emph{The Radon transform and some of its
  applications}}.
\newblock \bibinfo{publisher}{Courier Corporation}.
\newblock


\bibitem[Delaney and Bresler(1998)]%
        {Delaney1998}
\bibfield{author}{\bibinfo{person}{Alexander~H Delaney} {and}
  \bibinfo{person}{Yoram Bresler}.} \bibinfo{year}{1998}\natexlab{}.
\newblock \showarticletitle{Globally convergent edge-preserving regularized
  reconstruction: an application to limited-angle tomography}.
\newblock \bibinfo{journal}{\emph{IEEE Transactions on Image Processing}}
  \bibinfo{volume}{7}, \bibinfo{number}{2} (\bibinfo{year}{1998}),
  \bibinfo{pages}{204--221}.
\newblock


\bibitem[Faccio et~al\mbox{.}(2020)]%
        {Faccio2020non}
\bibfield{author}{\bibinfo{person}{Daniele Faccio}, \bibinfo{person}{Andreas
  Velten}, {and} \bibinfo{person}{Gordon Wetzstein}.}
  \bibinfo{year}{2020}\natexlab{}.
\newblock \showarticletitle{Non-line-of-sight imaging}.
\newblock \bibinfo{journal}{\emph{Nature Reviews Physics}} \bibinfo{volume}{2},
  \bibinfo{number}{6} (\bibinfo{year}{2020}), \bibinfo{pages}{318--327}.
\newblock


\bibitem[{Grau Chopite} et~al\mbox{.}(2020)]%
        {GrauCVPR2020}
\bibfield{author}{\bibinfo{person}{Javier {Grau Chopite}},
  \bibinfo{person}{Matthias~B. Hullin}, \bibinfo{person}{Michael Wand}, {and}
  \bibinfo{person}{Julian Iseringhausen}.} \bibinfo{year}{2020}\natexlab{}.
\newblock \showarticletitle{Deep Non-Line-of-Sight Reconstruction}. In
  \bibinfo{booktitle}{\emph{IEEE Conference on Computer Vision and Pattern
  Recognition (CVPR)}}.
\newblock


\bibitem[Guill{\'e}n et~al\mbox{.}(2020)]%
        {Guillen2020Effect}
\bibfield{author}{\bibinfo{person}{Ib{\'o}n Guill{\'e}n},
  \bibinfo{person}{Xiaochun Liu}, \bibinfo{person}{Andreas Velten},
  \bibinfo{person}{Diego Gutierrez}, {and} \bibinfo{person}{Adrian Jarabo}.}
  \bibinfo{year}{2020}\natexlab{}.
\newblock \showarticletitle{On the Effect of Reflectance on Phasor Field
  Non-Line-of-Sight Imaging}. In \bibinfo{booktitle}{\emph{IEEE International
  Conference on Acoustics, Speech and Signal Processing (ICASSP)}}. IEEE,
  \bibinfo{pages}{9269--9273}.
\newblock


\bibitem[Heide et~al\mbox{.}(2014)]%
        {Heide2014diffuse}
\bibfield{author}{\bibinfo{person}{Felix Heide}, \bibinfo{person}{Lei Xiao},
  \bibinfo{person}{Wolfgang Heidrich}, {and} \bibinfo{person}{Matthias~B
  Hullin}.} \bibinfo{year}{2014}\natexlab{}.
\newblock \showarticletitle{Diffuse mirrors: {3D} reconstruction from diffuse
  indirect illumination using inexpensive time-of-flight sensors}. In
  \bibinfo{booktitle}{\emph{IEEE Computer Vision and Pattern Recognition
  (CVPR)}}.
\newblock


\bibitem[Iseringhausen and Hullin(2020)]%
        {iseringhausen2020non}
\bibfield{author}{\bibinfo{person}{Julian Iseringhausen} {and}
  \bibinfo{person}{Matthias~B Hullin}.} \bibinfo{year}{2020}\natexlab{}.
\newblock \showarticletitle{Non-line-of-sight reconstruction using efficient
  transient rendering}.
\newblock \bibinfo{journal}{\emph{ACM Trans. Graph.}} \bibinfo{volume}{39},
  \bibinfo{number}{1} (\bibinfo{year}{2020}), \bibinfo{pages}{1--14}.
\newblock


\bibitem[Jarabo et~al\mbox{.}(2014)]%
        {Jarabo2014}
\bibfield{author}{\bibinfo{person}{Adrian Jarabo}, \bibinfo{person}{Julio
  Marco}, \bibinfo{person}{Adolfo Mu\~{n}oz}, \bibinfo{person}{Raul Buisan},
  \bibinfo{person}{Wojciech Jarosz}, {and} \bibinfo{person}{Diego Gutierrez}.}
  \bibinfo{year}{2014}\natexlab{}.
\newblock \showarticletitle{A Framework for Transient Rendering}.
\newblock \bibinfo{journal}{\emph{ACM Trans. Graph.}} \bibinfo{volume}{33},
  \bibinfo{number}{6} (\bibinfo{year}{2014}).
\newblock


\bibitem[Jarabo et~al\mbox{.}(2017)]%
        {jarabo2017recent}
\bibfield{author}{\bibinfo{person}{Adrian Jarabo}, \bibinfo{person}{Belen
  Masia}, \bibinfo{person}{Julio Marco}, {and} \bibinfo{person}{Diego
  Gutierrez}.} \bibinfo{year}{2017}\natexlab{}.
\newblock \showarticletitle{Recent advances in transient imaging: A computer
  graphics and vision perspective}.
\newblock \bibinfo{journal}{\emph{Visual Informatics}} \bibinfo{volume}{1},
  \bibinfo{number}{1} (\bibinfo{year}{2017}), \bibinfo{pages}{65--79}.
\newblock


\bibitem[Kadambi et~al\mbox{.}(2016)]%
        {kadambi2016occluded}
\bibfield{author}{\bibinfo{person}{Achuta Kadambi}, \bibinfo{person}{Hang
  Zhao}, \bibinfo{person}{Boxin Shi}, {and} \bibinfo{person}{Ramesh Raskar}.}
  \bibinfo{year}{2016}\natexlab{}.
\newblock \showarticletitle{Occluded imaging with time-of-flight sensors}.
\newblock \bibinfo{journal}{\emph{ACM Trans. Graph. (ToG)}}
  \bibinfo{volume}{35}, \bibinfo{number}{2} (\bibinfo{year}{2016}),
  \bibinfo{pages}{1--12}.
\newblock


\bibitem[Klein et~al\mbox{.}(2016)]%
        {Klein2016SR}
\bibfield{author}{\bibinfo{person}{Jonathan Klein}, \bibinfo{person}{Christoph
  Peters}, \bibinfo{person}{Jaime Mart{\'\i}n}, \bibinfo{person}{Martin
  Laurenzis}, {and} \bibinfo{person}{Matthias~B Hullin}.}
  \bibinfo{year}{2016}\natexlab{}.
\newblock \showarticletitle{Tracking objects outside the line of sight using
  {2D} intensity images}.
\newblock \bibinfo{journal}{\emph{Scientific Reports}}  \bibinfo{volume}{6}
  (\bibinfo{year}{2016}).
\newblock


\bibitem[La~Manna et~al\mbox{.}(2018)]%
        {LaManna2018error}
\bibfield{author}{\bibinfo{person}{Marco La~Manna}, \bibinfo{person}{Fiona
  Kine}, \bibinfo{person}{Eric Breitbach}, \bibinfo{person}{Jonathan Jackson},
  \bibinfo{person}{Talha Sultan}, {and} \bibinfo{person}{Andreas Velten}.}
  \bibinfo{year}{2018}\natexlab{}.
\newblock \showarticletitle{Error backprojection algorithms for
  non-line-of-sight imaging}.
\newblock \bibinfo{journal}{\emph{IEEE transactions on pattern analysis and
  machine intelligence}} \bibinfo{volume}{41}, \bibinfo{number}{7}
  (\bibinfo{year}{2018}), \bibinfo{pages}{1615--1626}.
\newblock


\bibitem[Laurenzis and Velten(2014)]%
        {Laurenzis2014feature}
\bibfield{author}{\bibinfo{person}{Martin Laurenzis} {and}
  \bibinfo{person}{Andreas Velten}.} \bibinfo{year}{2014}\natexlab{}.
\newblock \showarticletitle{Feature selection and back-projection algorithms
  for non-line-of-sight laser--gated viewing}.
\newblock \bibinfo{journal}{\emph{Journal of Electronic Imaging}}
  \bibinfo{volume}{23}, \bibinfo{number}{6} (\bibinfo{year}{2014}),
  \bibinfo{pages}{063003}.
\newblock


\bibitem[Liao et~al\mbox{.}(2021)]%
        {liao2021fpga}
\bibfield{author}{\bibinfo{person}{Zhengpeng Liao}, \bibinfo{person}{Deyang
  Jiang}, \bibinfo{person}{Xiaochun Liu}, \bibinfo{person}{Andreas Velten},
  \bibinfo{person}{Yajun Ha}, {and} \bibinfo{person}{Xin Lou}.}
  \bibinfo{year}{2021}\natexlab{}.
\newblock \showarticletitle{FPGA Accelerator for Real-Time Non-Line-of-Sight
  Imaging}.
\newblock \bibinfo{journal}{\emph{IEEE Transactions on Circuits and Systems I:
  Regular Papers}} (\bibinfo{year}{2021}).
\newblock


\bibitem[Lim et~al\mbox{.}(2015)]%
        {Lim2015}
\bibfield{author}{\bibinfo{person}{JooWon Lim}, \bibinfo{person}{KyeoReh Lee},
  \bibinfo{person}{Kyong~Hwan Jin}, \bibinfo{person}{Seungwoo Shin},
  \bibinfo{person}{SeoEun Lee}, \bibinfo{person}{YongKeun Park}, {and}
  \bibinfo{person}{Jong~Chul Ye}.} \bibinfo{year}{2015}\natexlab{}.
\newblock \showarticletitle{Comparative study of iterative reconstruction
  algorithms for missing cone problems in optical diffraction tomography}.
\newblock \bibinfo{journal}{\emph{Optics Express}} \bibinfo{volume}{23},
  \bibinfo{number}{13} (\bibinfo{date}{jun} \bibinfo{year}{2015}),
  \bibinfo{pages}{16933--16948}.
\newblock
\showISSN{1094-4087}
\href{https://doi.org/10.1364/OE.23.016933}{doi:\nolinkurl{10.1364/OE.23.016933}}


\bibitem[Lindell et~al\mbox{.}(2019)]%
        {Lindell2019wave}
\bibfield{author}{\bibinfo{person}{David~B Lindell}, \bibinfo{person}{Gordon
  Wetzstein}, {and} \bibinfo{person}{Matthew O'Toole}.}
  \bibinfo{year}{2019}\natexlab{}.
\newblock \showarticletitle{Wave-based non-line-of-sight imaging using fast fk
  migration}.
\newblock \bibinfo{journal}{\emph{ACM Trans. Graph.}} \bibinfo{volume}{38},
  \bibinfo{number}{4} (\bibinfo{year}{2019}), \bibinfo{pages}{1--13}.
\newblock


\bibitem[Liu et~al\mbox{.}(2019a)]%
        {Liu2019analysis}
\bibfield{author}{\bibinfo{person}{Xiaochun Liu}, \bibinfo{person}{Sebastian
  Bauer}, {and} \bibinfo{person}{Andreas Velten}.}
  \bibinfo{year}{2019}\natexlab{a}.
\newblock \showarticletitle{Analysis of feature visibility in non-line-of-sight
  measurements}. In \bibinfo{booktitle}{\emph{Proc. of the IEEE/CVF Conference
  on Computer Vision and Pattern Recognition}}. \bibinfo{pages}{10140--10148}.
\newblock


\bibitem[Liu et~al\mbox{.}(2020)]%
        {Liu2020phasor}
\bibfield{author}{\bibinfo{person}{Xiaochun Liu}, \bibinfo{person}{Sebastian
  Bauer}, {and} \bibinfo{person}{Andreas Velten}.}
  \bibinfo{year}{2020}\natexlab{}.
\newblock \showarticletitle{Phasor field diffraction based reconstruction for
  fast non-line-of-sight imaging systems}.
\newblock \bibinfo{journal}{\emph{Nat. Comm.}} \bibinfo{volume}{11},
  \bibinfo{number}{1} (\bibinfo{year}{2020}), \bibinfo{pages}{1--13}.
\newblock


\bibitem[Liu et~al\mbox{.}(2019b)]%
        {Liu2019phasor}
\bibfield{author}{\bibinfo{person}{Xiaochun Liu}, \bibinfo{person}{Ib{\'o}n
  Guill{\'e}n}, \bibinfo{person}{Marco La~Manna}, \bibinfo{person}{Ji~Hyun
  Nam}, \bibinfo{person}{Syed~Azer Reza}, \bibinfo{person}{Toan~Huu Le},
  \bibinfo{person}{Adrian Jarabo}, \bibinfo{person}{Diego Gutierrez}, {and}
  \bibinfo{person}{Andreas Velten}.} \bibinfo{year}{2019}\natexlab{b}.
\newblock \showarticletitle{Non-Line-of-Sight Imaging using Phasor Fields
  Virtual Wave Optics}.
\newblock \bibinfo{journal}{\emph{Nature}} (\bibinfo{year}{2019}).
\newblock


\bibitem[Liu and Velten(2020)]%
        {Liu2020ICCP}
\bibfield{author}{\bibinfo{person}{Xiaochun Liu} {and} \bibinfo{person}{Andreas
  Velten}.} \bibinfo{year}{2020}\natexlab{}.
\newblock \showarticletitle{The role of {Wigner} Distribution Function in
  Non-Line-of-Sight Imaging}. In \bibinfo{booktitle}{\emph{IEEE International
  Conference on Computational Photography (ICCP)}}.
\newblock


\bibitem[Maeda et~al\mbox{.}(2019)]%
        {maeda2019recent}
\bibfield{author}{\bibinfo{person}{Tomohiro Maeda}, \bibinfo{person}{Guy
  Satat}, \bibinfo{person}{Tristan Swedish}, \bibinfo{person}{Lagnojita Sinha},
  {and} \bibinfo{person}{Ramesh Raskar}.} \bibinfo{year}{2019}\natexlab{}.
\newblock \showarticletitle{Recent Advances in Imaging Around Corners}.
\newblock \bibinfo{journal}{\emph{arXiv preprint arXiv:1910.05613}}
  (\bibinfo{year}{2019}).
\newblock


\bibitem[Marco et~al\mbox{.}(2021)]%
        {Marco2021NLOSvLTM}
\bibfield{author}{\bibinfo{person}{Julio Marco}, \bibinfo{person}{Adrian
  Jarabo}, \bibinfo{person}{Ji~Hyun Nam}, \bibinfo{person}{Xiaochun Liu},
  \bibinfo{person}{Miguel Ángel Cosculluela}, \bibinfo{person}{Andreas
  Velten}, {and} \bibinfo{person}{Diego Gutierrez}.}
  \bibinfo{year}{2021}\natexlab{}.
\newblock \showarticletitle{Virtual light transport matrices for
  non-line-of-sight imaging}. In \bibinfo{booktitle}{\emph{2021 IEEE/CVF
  International Conference on Computer Vision (ICCV)}}.
\newblock


\bibitem[Margrave(2001)]%
        {Margrave2001}
\bibfield{author}{\bibinfo{person}{Gary~F. Margrave}.}
  \bibinfo{year}{2001}\natexlab{}.
\newblock \showarticletitle{Direct Fourier migration for vertical velocity
  variations}.
\newblock \bibinfo{journal}{\emph{GEOPHYSICS}}  \bibinfo{volume}{66}
  (\bibinfo{year}{2001}), \bibinfo{pages}{1504--1514}.
\newblock
Issue 5.


\bibitem[Margrave and Lamoureux(2019)]%
        {margrave_lamoureux_2019}
\bibfield{author}{\bibinfo{person}{Gary~F. Margrave} {and}
  \bibinfo{person}{Michael~P. Lamoureux}.} \bibinfo{year}{2019}\natexlab{}.
\newblock \bibinfo{booktitle}{\emph{Wave Propagation and Seismic Modeling}}.
\newblock \bibinfo{publisher}{Cambridge University Press}.
\newblock
\href{https://doi.org/10.1017/9781316756041.005}{doi:\nolinkurl{10.1017/9781316756041.005}}


\bibitem[Mertz(2019)]%
        {Mertz2019}
\bibfield{author}{\bibinfo{person}{Jerome Mertz}.}
  \bibinfo{year}{2019}\natexlab{}.
\newblock \bibinfo{booktitle}{\emph{Introduction to Optical Microscopy}
  (\bibinfo{edition}{2} ed.)}.
\newblock \bibinfo{publisher}{Cambridge University Press}.
\newblock


\bibitem[Moon(2014)]%
        {moon2014determination}
\bibfield{author}{\bibinfo{person}{Sunghwan Moon}.}
  \bibinfo{year}{2014}\natexlab{}.
\newblock \showarticletitle{On the determination of a function from an
  elliptical Radon transform}.
\newblock \bibinfo{journal}{\emph{J. Math. Anal. Appl.}} \bibinfo{volume}{416},
  \bibinfo{number}{2} (\bibinfo{year}{2014}), \bibinfo{pages}{724--734}.
\newblock


\bibitem[Nam et~al\mbox{.}(2021)]%
        {nam2021low}
\bibfield{author}{\bibinfo{person}{Ji~Hyun Nam}, \bibinfo{person}{Eric Brandt},
  \bibinfo{person}{Sebastian Bauer}, \bibinfo{person}{Xiaochun Liu},
  \bibinfo{person}{Marco Renna}, \bibinfo{person}{Alberto Tosi},
  \bibinfo{person}{Eftychios Sifakis}, {and} \bibinfo{person}{Andreas Velten}.}
  \bibinfo{year}{2021}\natexlab{}.
\newblock \showarticletitle{Low-latency time-of-flight non-line-of-sight
  imaging at 5 frames per second}.
\newblock \bibinfo{journal}{\emph{Nat. Comm.}} \bibinfo{volume}{12},
  \bibinfo{number}{1} (\bibinfo{year}{2021}), \bibinfo{pages}{1--10}.
\newblock


\bibitem[O’Toole et~al\mbox{.}(2018)]%
        {OToole2018confocal}
\bibfield{author}{\bibinfo{person}{Matthew O’Toole}, \bibinfo{person}{David~B
  Lindell}, {and} \bibinfo{person}{Gordon Wetzstein}.}
  \bibinfo{year}{2018}\natexlab{}.
\newblock \showarticletitle{Confocal non-line-of-sight imaging based on the
  light-cone transform}.
\newblock \bibinfo{journal}{\emph{Nature}} \bibinfo{volume}{555},
  \bibinfo{number}{7696} (\bibinfo{year}{2018}), \bibinfo{pages}{338}.
\newblock


\bibitem[Pueyo-Ciutad et~al\mbox{.}(2024)]%
        {pueyociutad2024polNLOS}
\bibfield{author}{\bibinfo{person}{Oscar Pueyo-Ciutad}, \bibinfo{person}{Julio
  Marco}, \bibinfo{person}{Stephane Schertzer}, \bibinfo{person}{Frank
  Christnacher}, \bibinfo{person}{Martin Laurenzis}, \bibinfo{person}{Diego
  Gutierrez}, {and} \bibinfo{person}{Albert Redo-Sanchez}.}
  \bibinfo{year}{2024}\natexlab{}.
\newblock \showarticletitle{{Time-Gated Polarization for Active
  Non-Line-of-Sight Imaging}}. In \bibinfo{booktitle}{\emph{Proc. of ACM
  SIGGRAPH Asia 2024}}.
\newblock


\bibitem[Rayleigh(1879)]%
        {Rayleigh1879}
\bibfield{author}{\bibinfo{person}{Lord Rayleigh}.}
  \bibinfo{year}{1879}\natexlab{}.
\newblock \showarticletitle{XXXI. Investigations in optics, with special
  reference to the spectroscope}.
\newblock \bibinfo{journal}{\emph{The London, Edinburgh, and Dublin
  Philosophical Magazine and Journal of Science}} \bibinfo{volume}{8},
  \bibinfo{number}{49} (\bibinfo{year}{1879}), \bibinfo{pages}{261--274}.
\newblock


\bibitem[Renna et~al\mbox{.}(2020)]%
        {Renna2020fast}
\bibfield{author}{\bibinfo{person}{Marco Renna}, \bibinfo{person}{Ji~Hyun Nam},
  \bibinfo{person}{Mauro Buttafava}, \bibinfo{person}{Federica Villa},
  \bibinfo{person}{Andreas Velten}, {and} \bibinfo{person}{Alberto Tosi}.}
  \bibinfo{year}{2020}\natexlab{}.
\newblock \showarticletitle{Fast-gated 16$\times$ 1 SPAD array for
  non-line-of-sight imaging applications}.
\newblock \bibinfo{journal}{\emph{Instruments}} \bibinfo{volume}{4},
  \bibinfo{number}{2} (\bibinfo{year}{2020}), \bibinfo{pages}{14}.
\newblock


\bibitem[Royo et~al\mbox{.}(2022)]%
        {royo2022non}
\bibfield{author}{\bibinfo{person}{Diego Royo}, \bibinfo{person}{Jorge
  Garc{\'\i}a}, \bibinfo{person}{Adolfo Mu{\~n}oz}, {and}
  \bibinfo{person}{Adrian Jarabo}.} \bibinfo{year}{2022}\natexlab{}.
\newblock \showarticletitle{Non-line-of-sight transient rendering}.
\newblock \bibinfo{journal}{\emph{Computers \& Graphics}}
  \bibinfo{volume}{107} (\bibinfo{year}{2022}), \bibinfo{pages}{84--92}.
\newblock
\href{https://doi.org/10.1016/j.cag.2022.07.003}{doi:\nolinkurl{10.1016/j.cag.2022.07.003}}


\bibitem[Royo et~al\mbox{.}(2023)]%
        {royo2023virtual}
\bibfield{author}{\bibinfo{person}{Diego Royo}, \bibinfo{person}{Talha Sultan},
  \bibinfo{person}{Adolfo Mu{\~n}oz}, \bibinfo{person}{Khadijeh
  Masumnia-Bisheh}, \bibinfo{person}{Eric Brandt}, \bibinfo{person}{Diego
  Gutierrez}, \bibinfo{person}{Andreas Velten}, {and} \bibinfo{person}{Julio
  Marco}.} \bibinfo{year}{2023}\natexlab{}.
\newblock \showarticletitle{Virtual Mirrors: Non-Line-of-Sight Imaging Beyond
  the Third Bounce}.
\newblock \bibinfo{journal}{\emph{ACM Trans. Graph.}} \bibinfo{volume}{42},
  \bibinfo{number}{4} (\bibinfo{year}{2023}).
\newblock
\href{https://doi.org/10.1145/3592429}{doi:\nolinkurl{10.1145/3592429}}


\bibitem[Saunders et~al\mbox{.}(2019)]%
        {saunders2019computational}
\bibfield{author}{\bibinfo{person}{Charles Saunders}, \bibinfo{person}{John
  Murray-Bruce}, {and} \bibinfo{person}{Vivek~K Goyal}.}
  \bibinfo{year}{2019}\natexlab{}.
\newblock \showarticletitle{Computational periscopy with an ordinary digital
  camera}.
\newblock \bibinfo{journal}{\emph{Nature}} \bibinfo{volume}{565},
  \bibinfo{number}{7740} (\bibinfo{year}{2019}), \bibinfo{pages}{472--475}.
\newblock


\bibitem[Seidel et~al\mbox{.}(2019)]%
        {seidel2019corner}
\bibfield{author}{\bibinfo{person}{Sheila~W Seidel}, \bibinfo{person}{Yanting
  Ma}, \bibinfo{person}{John Murray-Bruce}, \bibinfo{person}{Charles Saunders},
  \bibinfo{person}{William~T Freeman}, \bibinfo{person}{C~Yu Christopher},
  {and} \bibinfo{person}{Vivek~K Goyal}.} \bibinfo{year}{2019}\natexlab{}.
\newblock \showarticletitle{Corner occluder computational periscopy: Estimating
  a hidden scene from a single photograph}. In \bibinfo{booktitle}{\emph{2019
  IEEE International Conference on Computational Photography (ICCP)}}. IEEE,
  \bibinfo{pages}{1--9}.
\newblock


\bibitem[Tasinkevych and Trots(2014)]%
        {Tasinkevych2014circular}
\bibfield{author}{\bibinfo{person}{Jurij Tasinkevych} {and}
  \bibinfo{person}{Ihor Trots}.} \bibinfo{year}{2014}\natexlab{}.
\newblock \showarticletitle{Circular radon transform inversion technique in
  synthetic aperture ultrasound imaging: an ultrasound phantom evaluation}.
\newblock \bibinfo{journal}{\emph{Archives of Acoustics}} \bibinfo{volume}{39},
  \bibinfo{number}{4} (\bibinfo{year}{2014}), \bibinfo{pages}{569--582}.
\newblock


\bibitem[Torralba and Freeman(2012)]%
        {torralba2012accidental}
\bibfield{author}{\bibinfo{person}{Antonio Torralba} {and}
  \bibinfo{person}{William~T Freeman}.} \bibinfo{year}{2012}\natexlab{}.
\newblock \showarticletitle{Accidental pinhole and pinspeck cameras: Revealing
  the scene outside the picture}. In \bibinfo{booktitle}{\emph{2012 IEEE
  Conference on Computer Vision and Pattern Recognition}}. IEEE,
  \bibinfo{pages}{374--381}.
\newblock


\bibitem[Tsai et~al\mbox{.}(2017)]%
        {tsai2017geometry}
\bibfield{author}{\bibinfo{person}{Chia-Yin Tsai}, \bibinfo{person}{Kiriakos~N
  Kutulakos}, \bibinfo{person}{Srinivasa~G Narasimhan}, {and}
  \bibinfo{person}{Aswin~C Sankaranarayanan}.} \bibinfo{year}{2017}\natexlab{}.
\newblock \showarticletitle{The geometry of first-returning photons for
  non-line-of-sight imaging}. In \bibinfo{booktitle}{\emph{Proc. of the IEEE
  Conference on Computer Vision and Pattern Recognition}}.
  \bibinfo{pages}{7216--7224}.
\newblock


\bibitem[Velten et~al\mbox{.}(2012)]%
        {Velten2012nc}
\bibfield{author}{\bibinfo{person}{Andreas Velten}, \bibinfo{person}{Thomas
  Willwacher}, \bibinfo{person}{Otkrist Gupta}, \bibinfo{person}{Ashok
  Veeraraghavan}, \bibinfo{person}{Moungi~G. Bawendi}, {and}
  \bibinfo{person}{Ramesh Raskar}.} \bibinfo{year}{2012}\natexlab{}.
\newblock \showarticletitle{Recovering three-dimensional shape around a corner
  using ultrafast time-of-flight imaging}.
\newblock \bibinfo{journal}{\emph{Nat. Comm.}} \bibinfo{number}{3}
  (\bibinfo{year}{2012}).
\newblock


\bibitem[Wikipedia(2022)]%
        {wiki:Phased_array}
\bibfield{author}{\bibinfo{person}{Wikipedia}.}
  \bibinfo{year}{2022}\natexlab{}.
\newblock \bibinfo{title}{{Phased array} --- {W}ikipedia{,} The Free
  Encyclopedia}.
\newblock
\urldef\tempurl%
\url{http://en.wikipedia.org/w/index.php?title=Phased\%20array}
\showURL{%
\tempurl}
\newblock
\shownote{[Online; accessed 14-July-2022]}.


\bibitem[Xin et~al\mbox{.}(2019)]%
        {Xin2019theory}
\bibfield{author}{\bibinfo{person}{Shumian Xin}, \bibinfo{person}{Sotiris
  Nousias}, \bibinfo{person}{Kiriakos~N Kutulakos}, \bibinfo{person}{Aswin~C
  Sankaranarayanan}, \bibinfo{person}{Srinivasa~G Narasimhan}, {and}
  \bibinfo{person}{Ioannis Gkioulekas}.} \bibinfo{year}{2019}\natexlab{}.
\newblock \showarticletitle{A theory of {Fermat} paths for non-line-of-sight
  shape reconstruction}. In \bibinfo{booktitle}{\emph{IEEE Computer Vision and
  Pattern Recognition (CVPR)}}. \bibinfo{pages}{6800--6809}.
\newblock


\bibitem[Xu and Wang(2005)]%
        {Xu2005universal}
\bibfield{author}{\bibinfo{person}{Minghua Xu} {and} \bibinfo{person}{Lihong~V
  Wang}.} \bibinfo{year}{2005}\natexlab{}.
\newblock \showarticletitle{Universal back-projection algorithm for
  photoacoustic computed tomography}.
\newblock \bibinfo{journal}{\emph{Physical Review E}} \bibinfo{volume}{71},
  \bibinfo{number}{1} (\bibinfo{year}{2005}), \bibinfo{pages}{016706}.
\newblock


\bibitem[Yi et~al\mbox{.}(2021)]%
        {Yi2021SIGA}
\bibfield{author}{\bibinfo{person}{Shinyoung Yi}, \bibinfo{person}{Donggun
  Kim}, \bibinfo{person}{Kiseok Choi}, \bibinfo{person}{Adrian Jarabo},
  \bibinfo{person}{Diego Gutierrez}, {and} \bibinfo{person}{Min~H. Kim}.}
  \bibinfo{year}{2021}\natexlab{}.
\newblock \showarticletitle{Differentiable Transient Rendering}.
\newblock \bibinfo{journal}{\emph{ACM Trans. Graph. (Proc. SIGGRAPH Asia
  2021)}} \bibinfo{volume}{40}, \bibinfo{number}{6} (\bibinfo{year}{2021}).
\newblock


\bibitem[Young et~al\mbox{.}(2020)]%
        {Young2020Directional}
\bibfield{author}{\bibinfo{person}{Sean~I. Young}, \bibinfo{person}{David~B.
  Lindell}, \bibinfo{person}{Bernd Girod}, \bibinfo{person}{David Taubman},
  {and} \bibinfo{person}{Gordon Wetzstein}.} \bibinfo{year}{2020}\natexlab{}.
\newblock \showarticletitle{Non-line-of-sight Surface Reconstruction Using the
  Directional Light-cone Transform}. In \bibinfo{booktitle}{\emph{Proc. CVPR}}.
\newblock


\end{thebibliography}
